UC BERKELEY

CENTER FOR LONG-TERM CYBERSECURITY

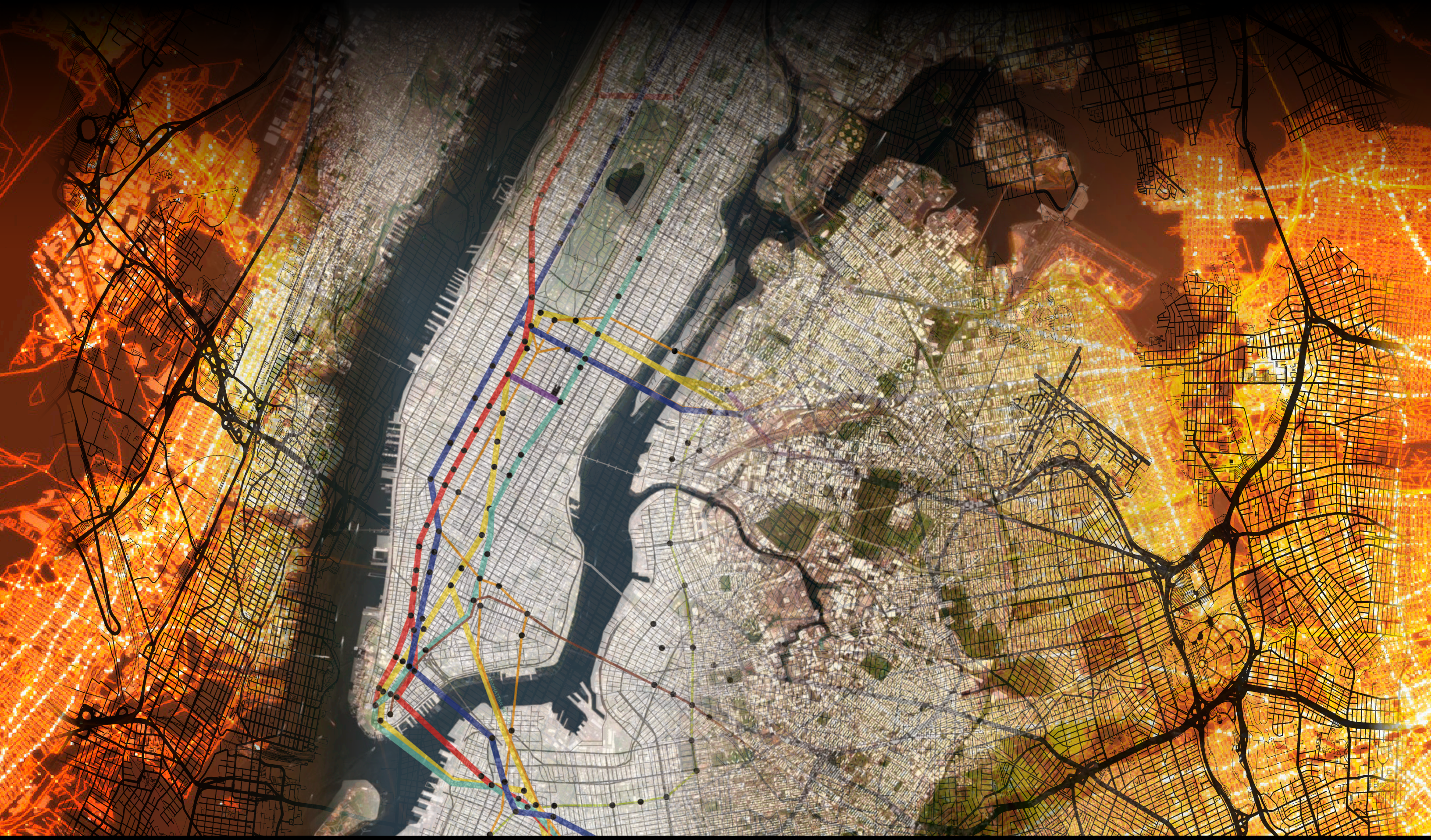

CLTC WHITE PAPER SERIES

# Choices, Risks, and Reward Reports

## Charting Public Policy for Reinforcement Learning Systems

THOMAS KRENDL GILBERT | SARAH DEAN | TOM ZICK | NATHAN LAMBERT

CLTC WHITE PAPER SERIES

# Choices, Risks, and Reward Reports

## Charting Public Policy for Reinforcement Learning Systems

THOMAS KRENDL GILBERT, SARAH DEAN, TOM ZICK, AND NATHAN LAMBERT

FEBRUARY 2022

CLTC
Center for Long-Term Cybersecurity
UC Berkeley

# Contents

# Executive Summary

Reinforcement learning (RL) is one of the most promising branches of foundational research in artificial intelligence (AI). RL formulates intelligence as a generic learning problem that, if solved, promises to match or exceed human capabilities. Current research and industrial applications of RL include social media recommendations, game-playing, traffic modeling, clinical health trials, and electric power utilities, among many other open-ended problems. In the long term, RL is considered by many AI theorists to be the most promising path to artificial general intelligence. This places RL practitioners in a position to design systems that have never existed before and lack prior documentation in law and policy.

This is both an exciting and frightening development. RL may make it possible to optimize how traffic moves through cities, how social media users consume and share content, and how health administrators oversee medical interventions, among many other applications. Public agencies could intervene on complex dynamics that were previously too opaque to deliberate about, and long-held policy ambitions would finally be made tractable. In this whitepaper we illustrate this potential and how it might be technically enacted in the domains of energy infrastructure, social media recommender systems, and transportation.

Alongside these unprecedented interventions come new forms of risk that exacerbate the harms already generated by standard machine learning tools. We correspondingly present a new typology of risks arising from RL design choices, falling under four categories: *scoping the horizon*, *defining rewards*, *pruning information*, and *training multiple agents*.

Beyond traditional machine learning (ML), there are two reasons why these risks constitute novel challenges for public policy. First, in ML, the primary risks have to do with *outputs* that a model generates (e.g., whether a model makes fair decisions). In RL, however, the risks come from the initial *specification of the task* (e.g., whether learning "good" behavior for automated vehicles also requires an initial definition of good traffic flow). Addressing these risks will require *ex ante* design considerations, rather than exclusively *ex post* evaluations of behavior.

Second, RL is widely used as a *metaphor* for replacing human judgment with automated decision-making. Even if end-to-end RL is not yet feasible, companies' strategic deployment of ML algorithms can be viewed as "human-driven RL." Whether the decision-maker is a machine, a human engineer, or a combination thereof, RL serves as a window into the wider forms of automation now pursued by technology firms and the risks those forms introduce.[1]

**Rather than allowing RL systems to unilaterally reshape human domains, policymakers need new mechanisms for the rule of reason, foreseeability, and interoperability that match the risks these systems pose.** Only in this way can design choices be structured according to terms that are well-framed, backed up by standards, and actionable in courts of law. We argue that criteria for these limits may be drawn from emerging subfields within antitrust, tort, and administrative law. It will then be possible for courts, federal and state agencies, and non-governmental organizations to play more active roles in RL specification and evaluation.

Applying RL to social domains requires designers to make technical choices, which may in turn change the way these domains work and how their dynamics can be managed. For example, to what extent can an RL agent nudge social media users to consume certain types of content, in order to increase platform engagement? Under what conditions could a self-driving car company's algorithm, for the sake of successful fleet operation, assume control of public roads without explicit permission or agreement? Could an RL-enabled electric utility appropriately respond to and optimize shifts in energy consumption over time? For designers to make sense of these questions, and to enable more active public policy deliberation, they need to document the choices they make when specifying a given system.

Building on the "model cards" and "datasheets" frameworks proposed by Mitchell et al.[2] and Gebru et al.[3], we argue the need for *Reward Reports* for AI systems. Reward Reports are living documents for proposed RL deployments that demarcate design choices. This includes which types of feedback have been brought into scope, what metrics have been considered to optimize performance, and why components of the specification (e.g., states, actions, rewards) were deemed appropriate. Reward Reports also outline the conditions for updating the report following *ex post* evaluation of system performance, the domain, or any interaction between these.

1 The Appendix further elaborates both these points in the context of emerging technical design paradigms for ML and RL.

2 Mitchell, Margaret, Simone Wu, Andrew Zaldivar, Parker Barnes, Lucy Vasserman, Ben Hutchinson, Elena Spitzer, Inioluwa Deborah Raji, and Timnit Gebru. "Model cards for model reporting." In *Proceedings of the conference on fairness, accountability, and transparency*, pp. 220–229. (2019).

3 Gebru, Timnit, Jamie Morgenstern, Briana Vecchione, Jennifer Wortman Vaughan, Hanna Wallach, Hal Daumé III, and Kate Crawford. "Datasheets for datasets." *arXiv preprint arXiv:1803.09010.* (2018).

A key recommendation from this whitepaper is that Reward Reports become a component of external oversight and continuous monitoring of RL algorithms running in safety-critical domains. Following recent work on the prospects of “foundation models”[4] and the need to document RL system capabilities,[5] we argue that Reward Reports are necessary for design decisions to be exhaustively evaluated. We also intend Reward Reports as a tool for AI governance beyond RL, enabling authorities to examine what a given system is optimizing for, and to evaluate the appropriateness of those terms. Reward Reports will ensure that design choices are auditable by third parties, contestable through litigation, and able to be affirmed by stakeholders.

We envision three key policy audiences for Reward Reports:

- Trade and commerce regulators
- Standards-setting agencies and departments
- Civil society organizations that evaluate unanticipated effects of AI systems

Our broader vision is that, in safety-critical domains, the role of an algorithm designer matures to be closer to that of a civil engineer: a technical expert whose credentials and demonstrated skills are trusted to oversee critical social infrastructure, and are worthy of certification by public authorities. Moreover, agencies such as the National Institute of Standards and Technology (NIST) and the Department of Transportation (DOT) must facilitate and guide critical design criteria so that the public interest remains in scope for advanced automated systems.

4 Bommasani, Rishi, Drew A. Hudson, Ehsan Adeli, Russ Altman, Simran Arora, Sydney von Arx, Michael S. Bernstein et al. “On the opportunities and risks of foundation models.” *arXiv preprint arXiv:2108.07258.* (2021).

5 Whittlestone, Jess, Kai Arulkumaran, and Matthew Crosby. “The Societal Implications of Deep Reinforcement Learning.” *Journal of Artificial Intelligence Research* 70 (2021): 1003–1030.

# Glossary

**Reinforcement learning:** A machine learning procedure based on exploring and taking actions sequentially to improve performance, without needing an initial definition of "good" behavior.

**Agent:** An autonomous entity that perceives, acts, and learns, directing its activity toward achieving goals.

**Environment (a.k.a. specification):** Every RL system consists of some agent that actively explores an *environment* that the designer has defined to be observable and navigable under certain conditions. These conditions make up the *specification* of the RL system, consisting of actions, rewards, observations, and states.

**Actions and behavioral policy:** The goal of an RL system is to design a good rule for choosing *actions* to take in the environment. This rule is called a *behavioral policy*. In the technical literature, it is simply referred to as a "policy," but we use the term "behavioral policy" to avoid overloading the term. For example, a self-driving car must choose steering angle and acceleration, and a content recommendation algorithm may suggest a video to particular users.

**Rewards:** The performance of a behavioral policy is judged by the *reward function,* which also serves as the directing force for learning (i.e., updating the policy). Rewards may be given by simple, direct measurements (e.g., "time spent watching a video" for a recommendation algorithm), a complex combination of measurements from the state and actions (e.g., lane position and fuel usage in autonomous vehicles), or even another learned *model* from a previously logged set of behaviors (e.g., an autonomous vehicle attempting to imitate previous trajectories). The system receives a reward signal at each timestep, and the sum of these individual rewards constitutes the cumulative reward (though there may be long periods of time where the system receives zero reward until completing some task). A common formulation in RL is to weigh near-term reward more heavily than long-term reward. This process, known as *discounting*, involves optimizing a sum of weighted rewards, where the importance of the reward at each time step is reduced by a constant factor known as the *discount rate*. For example, an RL-trained robot learning to run could be rewarded for every second it does not fall over, rather than for performing a specific sequence of actions, in order to encourage behavioral exploration.

**Observations and states:** An RL system takes actions on the basis of its *observations* about the environment, which may either be partial (not all features are visible) or total. Therefore, the behavioral policy is a map from the agent's *perceived state* to its actions. For example, a content recommendation system will choose videos (action) on the basis of a user's watch and interaction history (perceived state). In the common RL configuration known as a Markov Decision Process (MDP), the *state* summarizes all previous observations such that they are sufficient for predicting the future behavior of a system. For example, the future trajectory of a self-driving car can be determined by its current position and velocity, in addition to any forces (actions).

**Transition function (a.k.a. dynamics):** Within RL, an underlying system is often represented in a simpler form via a *transition function* that, given some action, denotes a probability of moving from one *state* to another. From the origins of the field in the area of optimal control, this function is often referred to as the *dynamics* of the system, even when it is only a partial approximation of a true system. The model often used to aid in simplifying transition functions is a Markov Decision Process, which makes the transition probabilities at any state independent of the past states, enabling RL to be a "memoryless" sequential decision-making problem. Throughout this paper, we will discuss how this simplification of dynamics and assumption of memorylessness can contribute to unintended risks in applied systems.

**Policy optimization:** At the heart of RL are the methods used to find an optimal behavioral policy (i.e., one that achieves the highest possible cumulative reward over some planning horizon). These methods can be both online, where an environment is explored or tested to update behavioral policies in real-time, or offline, where vast amounts of historical data are reweighted to design and estimate the effects of a new behavioral policy. Also, the methods can be *model-based*, meaning that they estimate a model of the transition function explicitly, or *model-free*, meaning that they optimize the policy without direct learning of the dynamics.

**Feedback:** In order for an RL system to learn a transition function and find an optimal behavioral policy, it must be extremely sensitive to information that appears when exploring its environment. This information could encourage the agent to favor certain sequences of actions over others, or to adopt new strategies for exploring, along with many other emergent effects. *Feedback* refers to the process of the agent drawing on sources of information from the current status of the environment as a guide to improve or modify its future behavior.

# Introduction

## WHAT IS REINFORCEMENT LEARNING?

The goal of reinforcement learning is to design a rule for choosing *actions* based on *observations* about the system. This rule is called a *behavioral policy*, and it is designed (or learned) by using data toward the goal of optimizing a cumulative *reward*. For example, by probing the characteristics and viewing history of users, an RL agent could learn what videos to recommend to particular people to maximize their time on a given website or application. This approach to learning is powerful, because it does not require *a priori* knowledge of domain dynamics (e.g., understanding how video recommendations will affect users' viewing behavior). RL agents learn and take actions over time, making them capable of procedural planning. However, an RL agent may learn a policy that is optimal with respect to how the environment was specified, but underdetermined with respect to the domain's purpose and real structure. For example, the goal of Twitter's users is presumably not to maximize time on site, but to add value to their lives by sharing and consuming meaningful content.

Bridging this gap is challenging, due to RL's distinctive technical affordances. Unlike more traditional machine learning approaches, RL algorithms are designed for a dynamic process of interactions, framed in terms of sequential rather than static decision-making. For example, in the context of content recommendations, RL methods make it possible to suggest videos in response to a sequence of interactions, rather than just the static interests of users. This has important implications both for the effectiveness of the system in achieving its goal (e.g., maximizing watch-time) and for the user experience (e.g., the quality or appropriateness of content).[6]

While RL can be directed to optimize any specified reward function, the RL paradigm itself is not value-neutral. In many cases, learning an optimal policy relies on "big data," along with the centralized technical infrastructure needed to process it. For robotics applications, these data are generated locally or through simulations. But in human-centered domains, the institutional paradigm of using big data to inform algorithms has been critically referred to as "surveillance capitalism,"[7] as it sacrifices the values of autonomy and privacy for the sake of profiting from

6 The technical details of sequential decision-making are discussed further in the Appendix.

7 Zuboff, Shoshana. *The Age of Surveillance Capitalism: The Fight for a Human Future at the New Frontier of Power*. New York: Profile Books (2019).

personal data. Even in cases where RL could demonstrably achieve a pre-specified goal, the specification may result in indirect costs for users or citizens, whose likely behaviors could be anticipated and shaped in advance of their expression and without their consent.

These risks are not speculative. For years, YouTube has come under fire for promoting disturbing children's content[8] and working as an engine of radicalization.[9] For example, a *New York Times* investigation found that videos promoted by YouTube's recommendations have "upended central elements of daily life" in Brazil, where far-right lawmaker Jair Bolsonaro was elected president in 2018. This all comes as YouTube achieved its goal of one billion hours of watch-time per day; a push on algorithm development was undoubtedly an important component of this success, with over 70% of views now coming from the recommended videos.[10] Examples from Facebook include an infamous study of "emotional contagion" among users,[11] the platform's role in Myanmar's ethnic strife,[12] recommender effects on political information-seeking and voting,[13] and an apparent mental health crisis among teenage users.[14] Though these social media sites largely rely on static machine learning adapted to sequential training data from watch history,[15] there has been a push within Google toward developing recommendation systems using RL.[16]

8 Bridle, James. *Something is Wrong on the Internet*. Vol. 6. Nov, 2017. URL: https://medium.com/@jamesbridle/something-is-wrong-on-the-internet-c39c471271d2

9 Nicas, Jack. "How YouTube Drives People to the Internet's Darkest Corners." *The Wall Street Journal*. February 7, 2018. URL: https://www.wsj.com/articles/how-youtube-drives-viewers-to-the-internets-darkest-corners-1518020478

10 Solsman, Joan E. "YouTube's AI is the puppet master over most of what you watch." URL: https://www.cnet.com/news/youtube-ces-2018-neal-mohan/ (2018).

11 Kramer, Adam DI, Jamie E. Guillory, and Jeffrey T. Hancock. "Experimental evidence of massive-scale emotional contagion through social networks." *Proceedings of the National Academy of Sciences* 111.24 (2014): 8788–8790.

12 Stevenson, Alexandra. "Facebook Admits It Was Used to Incite Violence in Myanmar." *New York Times*, November 6, 2018. URL: https://www.nytimes.com/2018/11/06/technology/myanmar-facebook.html

13 Bond, Robert M., et al. "A 61-million-person experiment in social influence and political mobilization." *Nature* 489.7415 (2012): 295–298.

14 Harwood, Graham. "When Even the SEC Can't Police Bad Behavior: The Facebook Whistleblower." *Chicago Policy Review* (*Online*) (2021).

15 Covington, Paul, Jay Adams, and Emre Sargin. "Deep neural networks for YouTube recommendations." In *Proceedings of the 10th ACM conference on recommender systems*, pp. 191–198. (2016).

16 See Ie, Eugene, Vihan Jain, Jing Wang, Sanmit Narvekar, Ritesh Agarwal, Rui Wu, Heng-Tze Cheng et al. "Reinforcement Learning for Slate-based Recommender Systems: A Tractable Decomposition and Practical Methodology." *arXiv preprint arXiv:1905.12767* (2019). See also Chen, Minmin, Alex Beutel, Paul Covington, Sagar Jain, Francois Belletti, and Ed H. Chi. "Top-K Off-Policy Correction for a REINFORCE Recommender System." In *Proceedings of the Twelfth ACM International Conference on Web Search and Data Mining*, pp. 456–464. (2019).

Academic research is only just beginning to grapple with the effects of platforms of this size.[17] And as RL applies even more naturally and effectively to sequential data, its implementation is poised to exacerbate current problems. Furthermore, it is not clear how to intervene to encourage better outcomes. Is political instability caused by the mere presence of conspiracy videos on Facebook, or by the company's decision to optimize a socially undesirable metric?

It is important not to lose sight of RL's potential strengths, particularly when applied to more conventional domains whose dynamics are well understood. In theory, RL could provide greater efficiency in the distribution of electrical power within energy grids. It could also fit efficient and safe traffic behaviors into corresponding road environments by optimizing the movements of citywide self-driving car fleets.[18] Assuming certain preconditions are met (i.e., important signals can be measured, sufficiently rich actions can be taken, and the dynamics are not too beholden to unmeasured phenomena), RL's application benefits are considerable. Long-held policy ambitions could be made actionable, and it will be possible to manage domains that were previously too complex or opaque. In brief, RL would help make it possible to choose how we want domains to work, including how service is rendered and to whom, and under what conditions.

This newfound agency should inspire policymakers' active engagement, rather than passive trepidation. In particular, algorithm designers need help specifying the goals and constraints of RL systems. If an RL agent's observations of its environment are insufficient to predict the future behavior of the system, this can generate unanticipated performance outcomes, and at worst total system failure. Furthermore, by defining the *state* and *observations*, the designers determine the scope of the system: are people part of the system's *state*? Are their behaviors included in the *observations*? Answers to these questions have important implications for privacy and behavioral manipulation. Social and digital media sites collect and exploit vast amounts of behavioral data, and tech companies are not held to established ethical standards for the human-subject experimentation implicit in any RL optimization.

Designing RL interventions requires understanding higher-level analytic distinctions that will pose new challenges for cybersecurity policy. For example, the *reward function* has been of significant interest to RL researchers as the motivating force of learning. Specifying rewards is a nontrivial but crucial task that often amounts to a decision of public policy. Oftentimes,

17 Ribeiro, Manoel Horta, Raphael Ottoni, Robert West, Virgílio AF Almeida, and Wagner Meira Jr. "Auditing Radicalization Pathways on YouTube." In *Proceedings of the 2020 Conference on Fairness, Accountability, and Transparency*, pp. 131–141. (2020).

18 Wu, Cathy, Aboudy Kreidieh, Kanaad Parvate, Eugene Vinitsky, and Alexandre M. Bayen. "Flow: Architecture and benchmarking for reinforcement learning in traffic control." *arXiv preprint arXiv:1710.05465* 10. (2017).

## CASE STUDY: VIDEO APP

To demonstrate the key aspects of reinforcement learning systems and how they work, we will consider a hypothetical video app. When a user opens the app, a video immediately begins to play. The user can swipe to skip to the next video. If they watch to the end of the video, the next video automatically plays. This hypothetical is a simplified version of apps that exist today, and it could be powered by a RL agent.

The RL agent determines how to show videos to a user when they open the app, skip, or reach the end of the previous video. Therefore, choosing the next video to play is the *action* for this agent. The app can base this decision on observations about the user: the videos they have viewed and skipped. The rule for selecting a video on the basis of these observations is the agent's *behavioral policy*.

First, let's consider a hypothetical policy. For the sake of simplicity, suppose that each video is about a single topic. Then the app could simply choose to show recent videos about the topic most commonly watched to completion by the user. This policy is hand-designed rather than learned. It clearly has some downsides: What if the user gets bored of seeing the same topic all the time? What about topics that the user would like, but that they haven't yet been shown?

Designing the policy with an RL algorithm instead can fix these problems, at least in principle. Instead of specifying a particular policy, we instead only need to determine a general structure: a function that maps the view history (a long array of timestamped video topics and watch percentage) to the next video. Notice that this encompasses policies that choose not to recommend the same topic many times in a row, potentially avoiding the problem of boredom.

Of course, it would be hard to determine the parameters of the policy by hand. The beauty of RL is that instead, we only have to identify a *reward signal*. A natural choice for this signal would be total watch-time. Once we have done this, the RL agent learns by doing: recommending videos and observing how much time the user watches. Based on these trials, the algorithm automatically updates the policy, iteratively improving it. A good RL algorithm trades off between exploration and exploitation, meaning that it avoids the problem of neglecting to show users topics that they might like.

the desired behavior is difficult to shape a *reward function* around (e.g., a robot folding laundry) or difficult to measure (e.g., enjoyment of a video). New technical methods, such as "inverse reward design,"[19] are meant to address problems of reward misspecification by designing systems to learn what designers want, not what designers say. While these approaches address some concerns, they do not fully explore the potential of design choices as a site of contestation, nor do they address the exogenous heuristics and constraints (and

19 Hadfield-Menell, Dylan, Smitha Milli, Pieter Abbeel, Stuart Russell, and Anca Dragan. "Inverse reward design." *arXiv preprint arXiv:1711.02827.* (2017).

corresponding normative assumptions) needed to select an appropriate reward function. Higher-level social environments (e.g., neighborhoods, social media profiles, cities, etc.) include competing interests that must somehow be reconciled, and not only on technical bases.

In what follows, we map the emergent political and economic stakes of RL specification onto the layers of feedback that are inherent to RL systems. This mapping illustrates the connections between distinct design choices and salient forms of normative risk, and also reveals corresponding public policy questions. Designers must be able to anticipate and communicate the likelihood of risks before they arise, and policymakers must be in a position to help adjudicate these choices so that they demonstrably reflect the public interest. In order to achieve this, we argue that designers will need to adopt the practice of *Reward Reports* that fully document the range of choices considered and ultimately made when the system was specified. Furthermore, governance recommendations should include Reward Reports as standard practice to ensure the safe development of advanced AI systems.

## HOW IS REINFORCEMENT LEARNING DISTINCT FROM ALTERNATIVE MACHINE LEARNING PARADIGMS?

Reinforcement learning is both more challenging to deploy and potentially more impactful than the other common frameworks for machine learning: unsupervised learning and supervised learning.

In unsupervised learning, the goal is *summarization* or *discovery* of information via learned representations or trends.[20] The system takes a large amount of data and creates some kind of summary of it. For example, you may take all of the videos someone has watched on YouTube and try to cluster them into categories. Maybe the categories will correspond to topic; maybe they will correspond to mood. The clustering method will take into account a variety of features about the video: how long it is, the channel it was posted to, its popularity, and so on. This approach can be thought of as "descriptive analytics" with a backwards-looking focus. Furthermore, the framework does not come with a predefined measure of performance, and does not judge whether one method of clustering videos is preferable to another. In general, the judgments are qualitative.

20 Our taxonomy here draws from Recht, Ben. "Make It Happen". Posted on January 29, 2018. Available online: http://www.argmin.net/2018/01/29/taxonomy/

In supervised learning, the goal is *prediction* via classification and regression analysis. Supervised learning comes with more structure in its framework. Rather than just aggregating a pile of data, supervised learning uses the features of each data point to predict the "label" assigned to it. For example, based on the length, content, and popularity of a given YouTube video — as measured by watch-time or "likes" or comments — the question is: can we predict how long a user will watch it? There is now a well-defined metric of performance: accuracy. This makes the object of concern a predictive model that corresponds to the metric of popularity.

In reinforcement learning, the goal shifts from summarization or prediction of passive observations to *navigation* of an environment. RL has a more precisely specified structure. Here, the data is not just a random collection of videos. Instead, the agent itself organizes the sequence of videos, recommendations, and watch-times. These correspond to *observations*, *actions*, and *rewards*. The object of concern is a behavioral policy, rather than a static representation chosen by the designer. This shift has two very important components. First, the focus is on selecting an action that maximizes the sum of future rewards, rather than just accurate prediction. What may have been a label in the supervised setting instead becomes the objective to maximize in the RL setting. And rather than merely make a prediction, the agent chooses actions to take. This moves us from predictive to prescriptive analytics. Second, reinforcement learning considers the maximization of reward by anticipating future consequences of current actions. As outlined further in the Appendix, reinforcement learning systems seek to solve *sequential decision-making* problems, which are repeated endeavors, rather than static and isolated actions. This changes the nature of prediction as we consider the longer-term effects of actions, or even the cumulative effects of particular sequences of actions. Rather than relying on passive observation as a heuristic for completing some task, the agent actively explores its environment based on whatever affordances (including observation) the designer has specified for it.

In practice, this shift means that the framework of RL reifies and technically enacts different types of feedback between learning algorithms, available sources of data, and the world. It does so explicitly at two different levels, and there is an additional level of feedback exterior to the formalism. Distinguishing these three types of feedback is crucial to understanding RL's transformative effects on the dynamics of particular domains, and the risks these effects can present. These points are outlined in Table 1 below, which summarizes the relationship between types of feedback, the corresponding information channel, the features at stake, and the effects on the dynamics of the environment.

| Type of feedback | Feedback channel | Dynamics | Specification element |
|---|---|---|---|
| Control | Agent-environment | Reaction | Actions |
| Behavioral | Explore-exploit | Evaluation | Rewards |
| Exogenous | Environment-domain | Drift | States, observations |

Table 1: The relationship between types of feedback, the channel through which information flows, the relationship to dynamics, and specification element(s).

The policy's act of controlling a system is the core type of feedback in RL. Policies constantly choose actions on the basis of the observations. Therefore, they create a feedback loop with the environment in a form of *control feedback*. A very simple example is a thermostat. Based on temperature sensors, thermostats decide whether or not to turn on a furnace. These "decisions" are made many times per second, and allow a heating/cooling system to keep a comfortable indoor temperature even under varying weather conditions. This type of feedback is often studied and designed as "automatic feedback control," and we will refer to it as **control feedback**. Other examples of control feedback are cruise control, where speedometer measurements are used to determine the acceleration of a car, or a dosing schedule for medications, which doctors follow to adjust dosages for patients based on their progress. "Intelligent" behaviors arise because the agent's actions are constantly adjusted on the basis of observations, even though the rules that control the behavior remain the same. This type of feedback underpins the function of RL, as is illustrated in Figure 1.

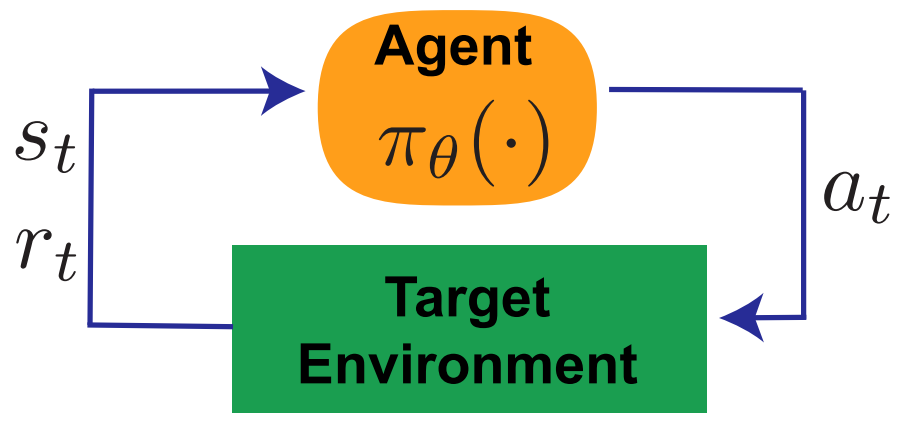

Figure 1: An illustration of control feedback showing the relationship between the agent and its environment, including a policy (pi) that maps actions (a) onto states (s) and rewards (r) according to policy parameters (theta).

There is a reason that such automatic feedback control systems are not usually called "reinforcement learning." It is because subject-matter experts (such as control engineers or medical specialists) are the designers of the policy parameters, rather than an algorithm based on those parameters. Control feedback takes different forms based on the learning algorithm. While supervised learning automatically generates the parameters that determine the performance of a *model*, RL automatically generates the parameters that determine the dynamics of a *behavioral policy*.

The RL framework holds that, although the training data is imperfect, the agent can find the data it needs by some form of exploration to improve the task. This is the second type of feedback present in RL systems: trial-and-error exploration of the environment, which we refer to as **behavioral feedback**. This is how the agent updates the policy, re-evaluating its past performance (which is recorded in system memory) in light of new actions taken. As an RL agent makes many sequential decisions, it receives evaluation from the environment in terms of reward, allowing it to decide whether to further explore the environment or exploit its current behavioral policy. Questions of *reaction* in control feedback become questions of trial-and-error *evaluation*: "At what temperature should the furnace turn on?" evolves into "What method would ensure the operating temperature is as close as possible to the target?" Reinforcement learning typically focuses on policies that maximize the cumulative reward. Although control engineers may also have a notion of reward or cost with which they validate their designs, reinforcement learning carries out the validation process systematically and automatically, and strives to do so optimally. The additional layer of behavioral feedback is reflected in Figure 2.

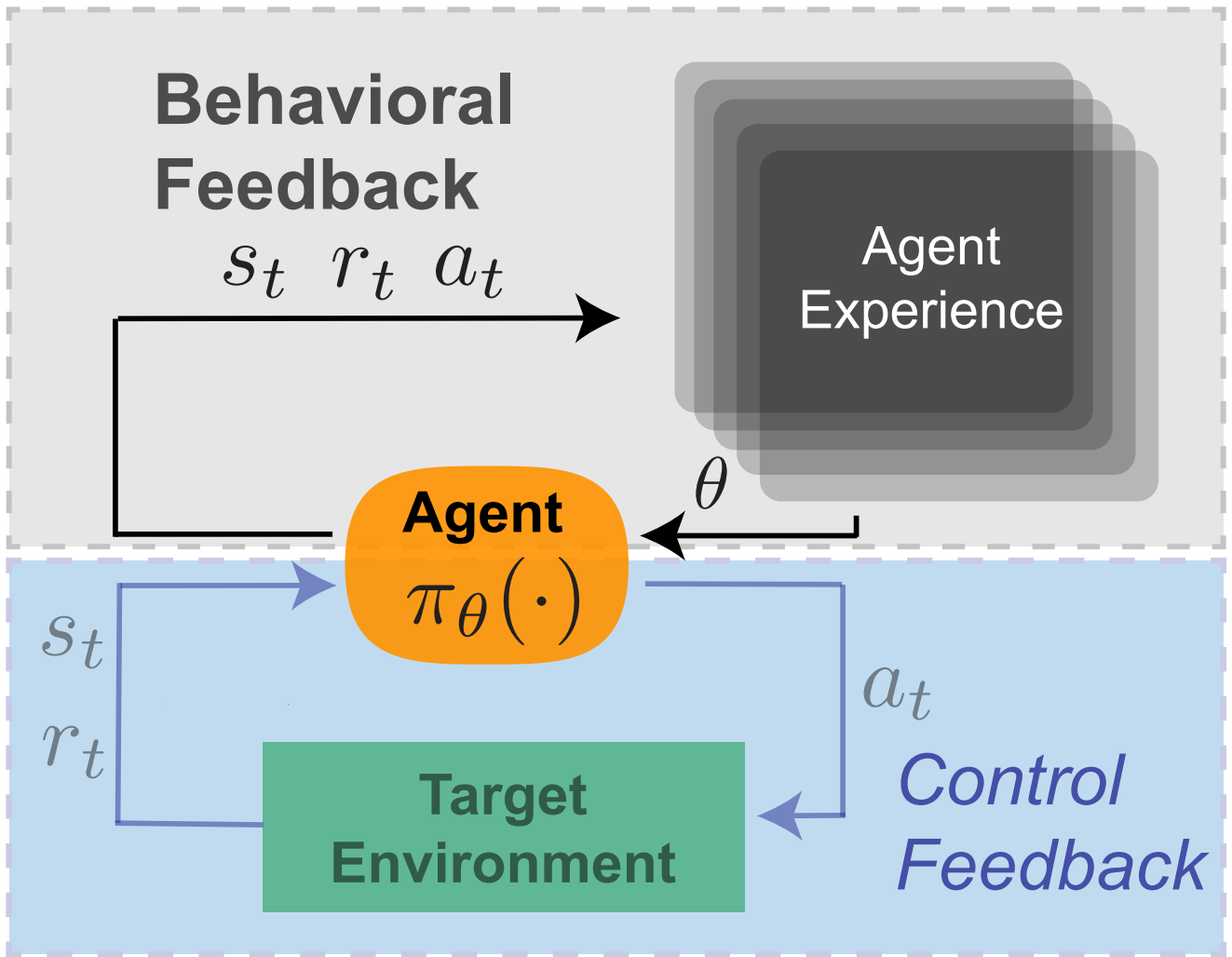

**Figure 2:** An illustration of behavioral feedback showing the relationship between the agent and its own replay memory, from which sequential actions are incorporated into behavior (theta).

In particular, the reinforcement learning methods that most often capture the imagination are those that are updated online: on the basis of operational experience, the parameters of the policy are updated through trial and error. At each instant, an RL agent is reactive to the environment through control feedback; over time, it adjusts that policy based on its performance. For example, by evaluating how well the temperature is regulated within a room, a smart thermostat could adjust how quickly it turns on a furnace to keep the temperature more steady in the future.

This "learning from experience" is part of what makes RL systems seem so powerful, and so applicable to domains that are difficult to otherwise model. Control engineers have had great success with chemical plants, transportation, HVAC, and other physical systems, but they would be challenged to hand-design policies for recommending music to a listener. Many human domains lack physical models and are much better approached via data, and therefore the framework of RL opens them up to policy design. However, behavioral feedback also makes RL systems uniquely challenging to govern, as it is difficult to establish responsibility for ongoing "training" of the system in user-specified conditions. If someone starts watching violent extremist content on YouTube, and the system then learns to offer more of that content, who is responsible for that person now only receiving violent extremist recommendations?[21]

The final type of feedback is exogenous or **exo-feedback**, in which the domain itself shifts in response to the deployed RL policy. These shifts could be due to political or economic conditions that are outside the purview of the RL algorithm, but interact with it in ways that may cause the environment to *drift* over time. For example, smart thermostats may enable finer-grained control over household heating, changing what a comfortable home amounts to or changing the electric loading of buildings, towns, and regions. Video recommendations given by a recommender system might incentivize YouTube creators to create attention-grabbing content, turning to strategies like outrage and conspiracy. As the breadth and duration of reinforcement learning systems increases, the potential for interaction between the learning environment and application domain — or degree of variation across users — continues to grow. This mode of feedback is presented in Figure 3 as an addition to a traditional rendition of the RL system, capturing the potential of externalized risks of this framework.

These forces are not unique to RL. If models are retrained on fresh data, the naive application of supervised predictions can also exhibit behavioral feedback as well as exo-feedback. This

21 We thank Jared Brown for this point and example.

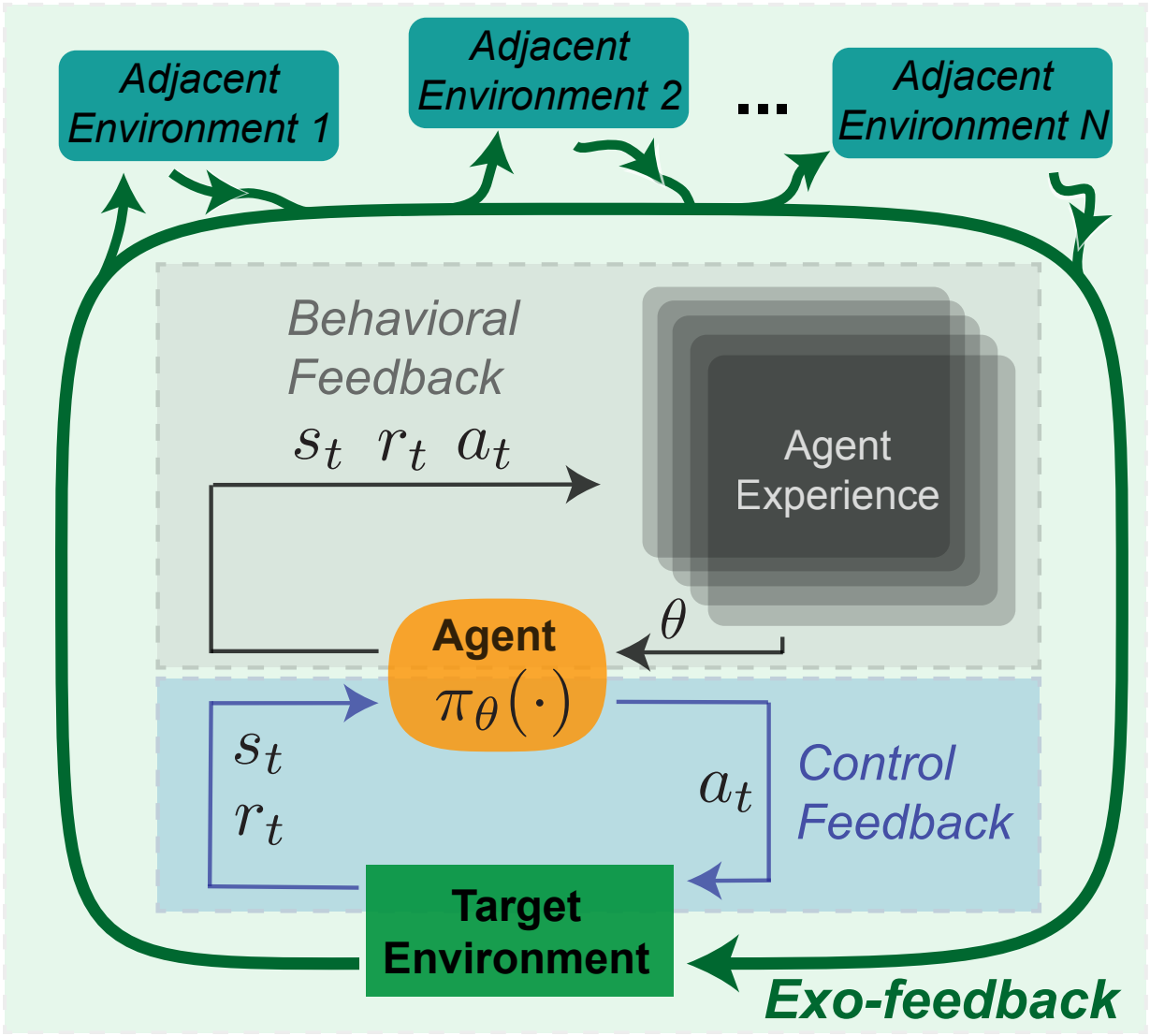

**Figure 3:** An illustration of exo-feedback in which control and behavioral feedback interacts with other parts of the application domain, causing the environment to drift over time.

problem is now explored by the growing field of performative prediction.[22] RL also has its own technical challenges, including the difficulty of training agents to meet standard performance benchmarks in even highly-constrained simulation environments.[23]

Still, there are two reasons why RL requires direct attention from crafters of public policy. First, RL serves as a *technical paradigm* that incorporates the notion of behavioral feedback directly in the chosen optimization, resulting in specifications that carry types of risks directly in their formal assumptions. As a result, the technical design risks we focus on in this paper should be understood as *ex ante* design considerations, rather than exclusively *ex post* evaluations, as is common for other iterative machine learning solutions. Second, RL is trusted by practitioners as a *metaphor* for automating human-in-the-loop components of current machine learning pipelines. This makes RL, regardless of its technical viability, a window into the risk-laden automation capacities of technology firms, as well as the particular domains in which those risks are most likely to emerge and what governance mechanisms will be needed to respond to them.

22 Perdomo, Juan, Tijana Zrnic, Celestine Mendler-Dünner, and Moritz Hardt. "Performative prediction." In *International Conference on Machine Learning*, pp. 7599–7609. PMLR, (2020).

23 See for example the following papers: Islam, Riashat, Peter Henderson, Maziar Gomrokchi, and Doina Precup. "Reproducibility of benchmarked deep reinforcement learning tasks for continuous control." *arXiv preprint arXiv:1708.04133*. (2017). Henderson, Peter, Riashat Islam, Philip Bachman, Joelle Pineau, Doina Precup, and David Meger. "Deep reinforcement learning that matters." In *Proceedings of the AAAI conference on artificial intelligence*, vol. 32, no. 1. (2018). Mania, Horia, Aurelia Guy, and Benjamin Recht. "Simple random search provides a competitive approach to reinforcement learning." *arXiv preprint arXiv:1803.07055*. (2018). Agarwal, Rishabh, Max Schwarzer, Pablo Samuel Castro, Aaron Courville, and Marc G. Bellemare. "Deep reinforcement learning at the edge of the statistical precipice." *arXiv preprint arXiv:2108.13264*. (2021).

# A Typology of Choices and Risks in Reinforcement Learning Design

The capabilities of RL systems permit novel strategies for optimal performance. They also open the application domain to dangers that must be mitigated. In this section, we discuss four types of RL design choices, highlighting how choices about the specification can generate particular socio-technical risks for which designers must plan. These choices are briefly defined here:

- **Scoping the Horizon:** The specification of some exploratory limit against which an RL agent learns to perform an arbitrarily simple or complex task.
- **Defining Rewards:** The specification of the reward pursued by an RL agent as it seeks to incorporate a behavioral policy within some environment.
- **Pruning Information:** The specification of actions and states so that an RL agent can learn a policy more efficiently or reliably according to a given task.
- **Training Multiple Agents:** The specification of an observable environment with agents that pursue a goal either collectively, collaboratively, competitively, or in parallel.

The risks arising from these choices are further delineated in Table 2.

| Design choice | Specification element(s) | Type(s) of feedback | Resultant risk |
|---|---|---|---|
| Scoping the Horizon | action space, policy | all | Inappropriate Flow |
| Defining Rewards | reward | behavioral | Reward Hacking |
| Pruning Information | actions, states | control | Regulatory Capture |
| Training Multiple Agents | observations | exogenous | Goodhart's Law |

**Table 2:** The relationships between RL design choices, feedback, and risks.

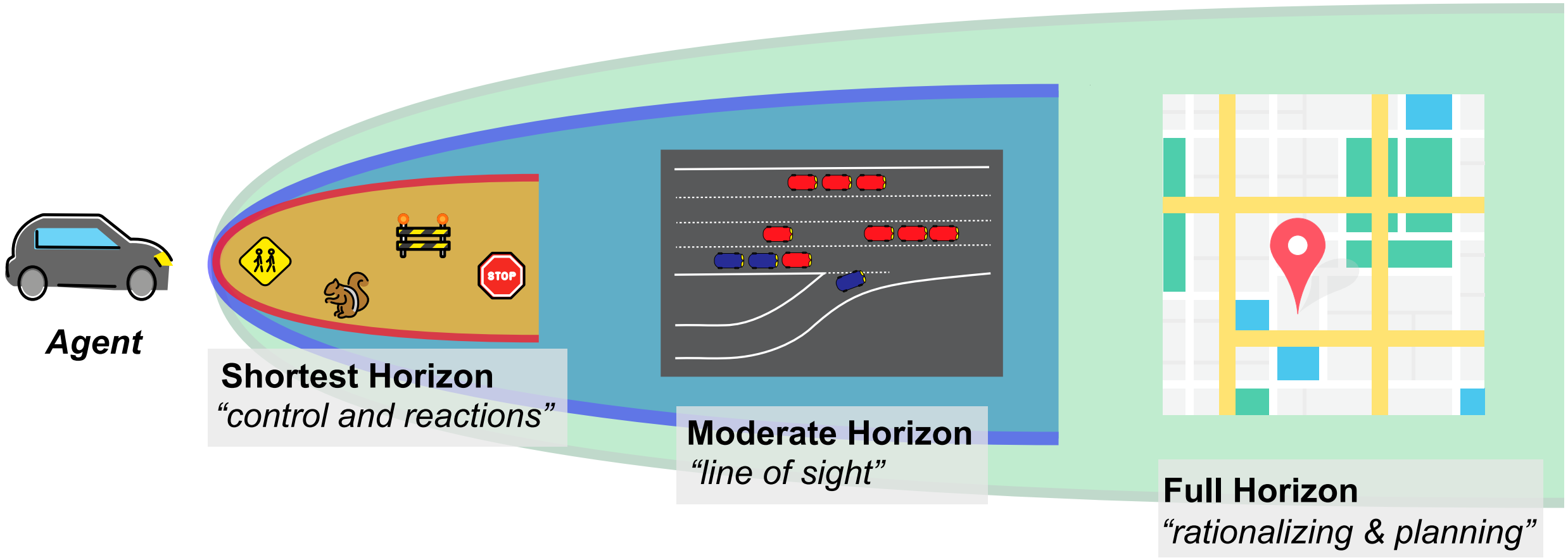

**Figure 4:** Distinct planning horizons for vehicle behavior. A short horizon comprises immediate reactions to nearby objects (e.g., signage, road obstacles). A longer horizon comprises more strategic, line-of-sight behaviors (e.g., merging, signaling, passing). Even longer horizons could capture end-to-end route planning. As the horizon expands, different dynamics are brought into scope.

## SCOPING THE HORIZON

Paramount to RL is the horizon against which the agent performs actions. The agent itself does not decide which dynamics are internal to that horizon (i.e., able to be learned) and which are external. Instead, the designer must choose — often without precedent — how far the agent is able to "see" and plan on its own, as well as the dynamics it is able to incorporate as a basis for its behavior. As illustrated in Figure 4, these choices may require translating complex human activities and domains into discrete states, actions, and rewards. This translation must be strictly tested, as even a robust policy may catastrophically fail when confronted with edge cases.

Consider the case of an autonomous vehicle (AV) that is learning how to avoid hitting road obstacles. Make the planning horizon too short, and the AV may learn to slam on the brakes immediately before reaching a stop sign. This not only wears down the AVs' braking system components, but would also make other drivers react defensively, perhaps making the system less safe. Make the planning horizon too long, and the AV may learn to drive well below the speed limit in order to conserve fuel or prepare for congested intersections miles ahead. This may lead to road rage, suboptimal traffic coordination, or accidents related to human factors. The point is that human behaviors include a mix of automatic responses and active attention, and complex activities like driving are made up of many behaviors at multiple layers of abstraction. A related term from computer science is *compositionality*: the way that the behaviors of a given agent are similar in form to the rules that all the agents are following at a given moment.[24]

## Inappropriate Flow

A longer or shorter planning horizon is not inherently good or bad. Rather, it is a matter of the nature of the activity the RL agent is trying to learn, on what abstraction layer it operates, and what kinds of tasks that layer comprises. RL practitioners may argue that a longer horizon is often better, as the agent learns to incorporate richer forms of behavioral feedback without the designer having to re-specify to a particular abstraction layer. In fact, recent research on recommendation systems has tried to incorporate the satisfaction of content providers by using a longer planning horizon, making the recommender sensitive to what keeps both providers and users on (or off) the platform.[25] Hence, the agent becomes "smarter" by being given a bigger set of "eyes" with which to see content producers and consumers.

But one of the key arguments of this whitepaper is that behavioral feedback cannot be "naturally" incorporated within an RL specification. To expand the planning horizon, information must flow in ways that will permit optimization at unprecedented scales. Expanding or contracting the horizon opens or closes the system to other types of control or exo-feedback that may also comprise the domain but never have been cohesively organized in terms of behavior. Moreover, domains do not exist in isolation from each other; humans learn to navigate between as well as within them, following a complex web of norms, laws, and protocols that may lack explicit documentation. While typically far from consideration in RL specification, there is a real danger that behaving more and more optimally will disintegrate these factors in ways that are difficult to trace. If these issues are ignored, the designer risks permitting inappropriate flow of information between stakeholder roles.[26]

The designer's decision to expand or contract the horizon of a given RL specification must be further documented in light of risks we describe in the following sections, in connection with

24 Ghani, Neil, Jules Hedges, Viktor Winschel, and Philipp Zahn. "Compositional game theory." In *Proceedings of the 33rd Annual ACM/IEEE Symposium on Logic in Computer Science*, pp. 472–481. (2018).

25 Zhan, Ruohan, Konstantina Christakopoulou, Ya Le, Jayden Ooi, Martin Mladenov, Alex Beutel, Craig Boutilier, Ed Chi, and Minmin Chen. "Towards Content Provider Aware Recommender Systems: A Simulation Study on the Interplay between User and Provider Utilities." In *Proceedings of the Web Conference 2021*, pp. 3872–3883. (2021).

26 Nissenbaum, Helen. *Privacy in context*. Stanford University Press, 2009. Drawing from Nissenbaum and other researchers of contextual integrity, we note that the inappropriate flow of information transcends privacy concerns and may affect a range of other human values and concerns. We defer a more complete analysis of this topic to future work.

other choices about the specification. This documentation is particularly needed to consider "tail-risk" situations in which an agent's behavior, while typically optimal, turns out to be brittle and potentially catastrophic. This can arise due to an agent being inadequately or excessively sensitive to domain dynamics, arising from contingencies in how the agent was trained. Consider an autonomous vehicle operating on a standard four-lane highway versus a complex intersection in Manhattan: how can the designer account for and demonstrate control over behaviors across these two very different settings? And how can this documentation anticipate legitimate concerns about inappropriate flow?

In either case, third-party regulators would need complete documentation of the proposed behavior policy, the specification behind how the model was learned, and other possible setups that were considered, but rejected by the designer. Only by doing this can systems be certified as having undergone a process of documentation that accounts for the design choices behind a given learned model. It follows that, regardless of application, broader planning horizons will require domain-specific standards of accountability, oversight, and responsibility so that after-effects of the optimization do not "spill over" to harm the integrity of the domain at stake. RL designers will need specialized training, access to data protocols, and liability constraints in their toolbox so that modeling choices reflect appropriate attention to informational integrity.

## DEFINING REWARDS

Defining rewards, referred to in the technical literature as "reward designing," means that the designer alters what counts as "reward" within the RL system and observes the effects on the agent's behavior. Rewards are most often redefined for two reasons. First, it is often hard to define what intelligence means in terms of behavior, and designers may resort to proxy definitions based on what can be measured. For example, there is no perfect definition of driving, but we all agree that hitting things is bad, so automated vehicles can at least be rewarded for not getting into accidents. Second, many human activities are defined by what designers call "sparse rewards," meaning that positive feedback is extremely infrequent. To continue with the above example, the agent may be rewarded for not causing an accident only after finishing an entire trip, with no signal encouraging the agent along the way. Think of how hard it would be to know whether you are understanding class material if your only graded assignment is the final exam, and you can imagine how hard it is for AVs to drive through even a simple neighborhood, let alone an entire city grid, without rewards to guide them.

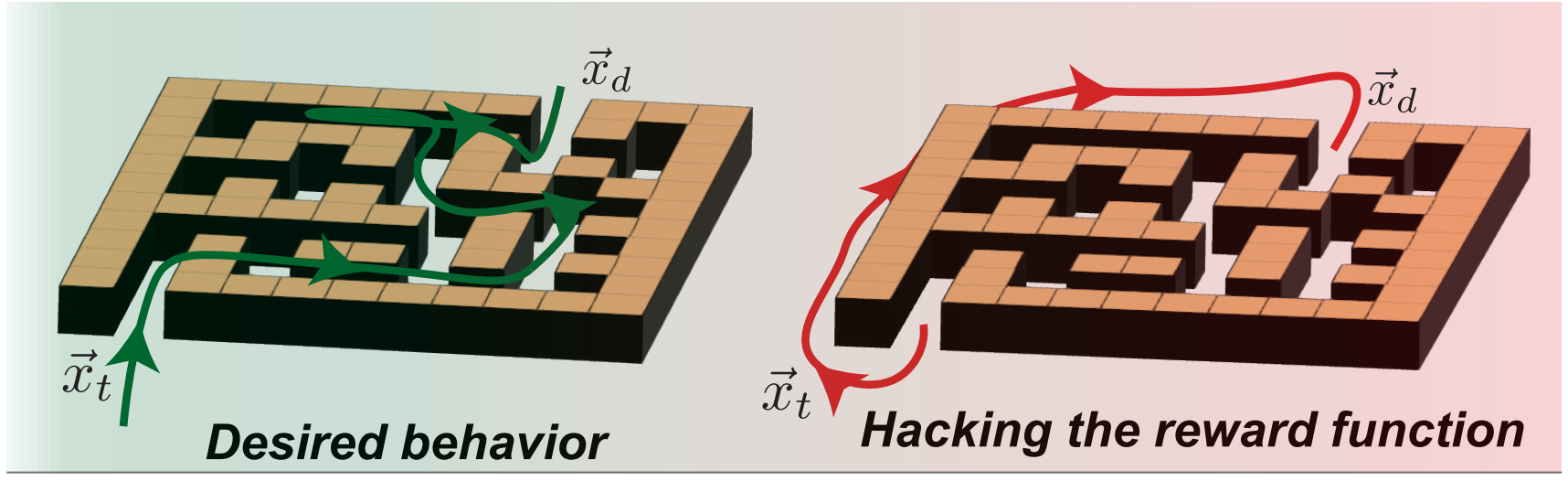

$$r(s_t, a_t) = -\|\vec{x}_t - \vec{x}_d\|^2$$

(Reward is a form of "*Minimize distance to goal*")

Figure 5: Defining rewards can lead to reward hacking if the agent learns to navigate around a maze rather than through it.

## Reward Hacking

Redefining rewards may make it easier for the agent to learn complicated tasks,[27] decompose an intractable activity into discretized task modules, or simply be of theoretical interest to the designer. This might also be done to test whether an optimal policy could be learned for more complex environments. Whatever the motivation, (re)defining rewards assumes that there is a single optimal behavior to be learned. This means that the agent could learn a policy that precisely accords with how the designer has redefined rewards, generating behaviors that conform to the "letter" of the environment but not the "spirit" of the activity. This results in what technical researchers call *reward hacking*, or more colloquially the "King Midas problem," when agents that do precisely what they were asked in a way that no one expects or wants. An example of reward hacking based on maze navigation is illustrated in Figure 5.

While defining rewards is a necessary component of good RL design, reward hacking is a reminder for the designer to be specific about *how* their desired outcome is meant to be accomplished. Reward design assumes that the state-action space of the environment is temporally stable and conforms to the real dynamics of the domain, or can be adjusted with no significant change of meaning. But this is a problem in environments whose rewards are either difficult to specify or inconsistent, that are populated by multiple agents with possibly distinct goals, or that lack consensus on what counts as good behavior. Even if the designer has successfully optimized control and behavioral feedback, reward hacking could generate exo-feedback that is unstable or undesirable. Other parts of the environment — including humans — would then interact with the RL system in uncontrolled or unpredictable ways.

27 See, for example, Ecoffet, Adrien, Joost Huizinga, Joel Lehman, Kenneth O. Stanley, and Jeff Clune. "Go-explore: a new approach for hard-exploration problems." *arXiv preprint arXiv:1901.10995*. (2019).

Designers are not working completely in the dark. It is possible to conduct "ablation studies" on the reward function, selectively removing system components or proxy metrics to observe the resulting effect on the behavioral policy.[28] This makes it technically easier to determine if specific metrics are being "hacked" or playing an outsized role in the learned policy. But guess-and-check ablation studies may still leave multiple reward definitions on the table that, while technically viable, reflect different interpretations of the designer's role vis-a-vis the application domain. Consider the example of training an RL agent to wash dirty dishes. When ablating rewards, the designer must be able to answer the question: "Is what I am doing more like designing a dishwasher, or teaching a child to wash dishes?" The former is akin to supplying a proxy metric for "clean dish" that the agent learns from scratch, as a dishwasher itself is a kind of proxy for doing the dishes (since a dishwasher does not understand what a dish is, why it is washing it, or what happens to it afterward). This proxy definition may then feed back onto the activity of dishwashing, implicitly redefining the use and function of the sink, hand towels, and counter space. However, teaching the agent merely provides more constant and reliable feedback for it to observe and interact with in real time. In this case, the activity itself is untouched, as the designer is delegating some pre-specified task for an artificial agent to complete. The difference lies in the ambiguous role of the designer as either someone with the power to actively reorganize how information flows through the domain (a kind of major-domo), or as a passive source of information for how a given task is conducted (a kind of teacher).

The problem is not only that a private company could assume ownership of a domain and unilaterally decide what to optimize for. The deeper issue is that a bound must be set on what type(s) of dynamics can be learned, and that bound may be easier to specify for some domains than others. The basis for this may be either natural (e.g., rooted in the physicality of the environment) or artificial (choices were made about how to define and structure salient activities). Much of RL research already focuses on learning probability distributions that cannot be predicted with certainty but still clear desired thresholds of performance, based on assumptions about the data structure, choice of learnable features, and computation available for modeling. But in order for reward shaping to avoid dysfunctionally manifesting as reward hacking, this exclusively technical work must be enriched and informed by appropriate specification of the structure in which actions may be undertaken.

For example, ablations would need to follow a documented experimental procedure that establishes the grounds for their insights and why the resulting behavior can be trusted. This

28 Hessel, Matteo, Joseph Modayil, Hado Van Hasselt, Tom Schaul, Georg Ostrovski, Will Dabney, Dan Horgan, Bilal Piot, Mohammad Azar, and David Silver. "Rainbow: Combining improvements in deep reinforcement learning." In *Thirty-second AAAI conference on artificial intelligence*. (2018).

may include specifying the planning horizon against which the ablation is performed, the metrics included or rejected, and the reasons for both, in light of what is known about the domain. Put differently, the point of defining rewards is to determine how to make behavioral feedback both possible (by choosing to optimize something measurable) and effective (by defining the reward signal to be easily learnable), without compromising the integrity of the domain. It follows that good RL documentation must include rich measurements of agent behavior not included in the reward function, as well as make sense of the reward definition in light of the overall objective.[29]

## PRUNING INFORMATION

Designers must also structure the environment so that the agent can observe the reward signal or states, however they have been defined, only under precise conditions. Here the issue is not sparsity but abundance, as the designer delimits the number of ways the agent can learn to match observed rewards with actions taken. Consider the difference between monitoring the eye movements of curbside pedestrians to predict intent to enter the crosswalk, and simply maintaining a standard distance behind all other road users. The latter prunes information to more crudely approximate "safe" driving behavior in a traffic setting, but is also many orders of magnitude easier to train. Information pruning could be motivated by finite computation or limited dataflow, or a designer may want to test the agent's ability to navigate complex environments using arbitrarily simple signals. The agent may learn more reliably or efficiently, even in environments with very large state-action spaces. But in environments with diverse sources of information (as in the case of visually-focused autonomous navigation highlighted in Figure 6), this pruning may compel the designer to prioritize some sensory inputs at the expense of others in pursuit of learning an optimal policy.

A good example from supervised learning is Tesla's decision to pursue a "camera only" policy for its Autopilot software.[30] Whereas Waymo uses a massive suite of many kinds of vehicle sensors (including cameras, multiple sources of LiDAR, and radar), Tesla simplified the types and structure of data needed for its software to operate by committing to a single perception mode, removing latency issues across sensors. The company claimed that Autopilot cameras were at least as good as human eyesight, with simpler and more reliable engineering. However, this design choice was not accompanied by documentation of how the change affected

29 We thank Dylan Hadfield-Menell for this point.

30 Fallon, Patrick and Juliette Michel. "Tesla's cameras-only autonomous system stirs controversy." Techxplore. January 6, 2022. https://techxplore.com/news/2022-01-tesla-cameras-only-autonomous-controversy.html

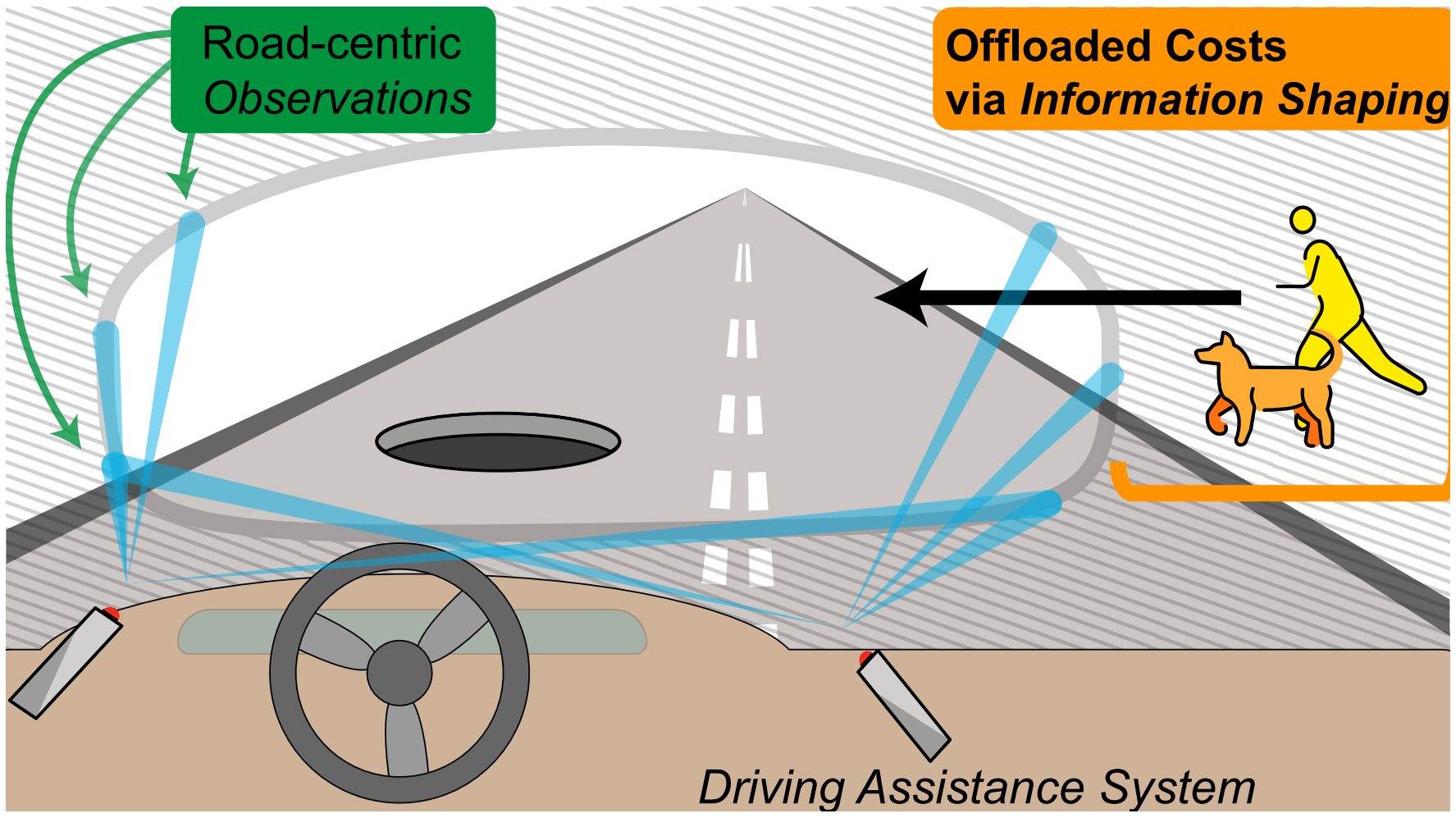

Figure 6: Information pruning in the context of traffic motion planning. The system includes actions and states only on the road itself, ignoring more costly features (e.g., pedestrians).

vehicle behaviors (e.g., following distance), reshaped values specific to the domain (e.g., safety, liability, mobility), or saved costs (either computational or financial). Absent corresponding documentation, the design choice cannot be evaluated for its positive or negative effects for Tesla owners or other road users.

## Regulatory Capture

This case demonstrates how information pruning intentionally simplifies control feedback in order to make it easier to learn from behavioral feedback. In effect, it normalizes some expected behaviors while underspecifying other available modes of communication. Pruning gives the designer the discretion to delimit what interface(s) matters most in coordination problems between different types of agents. In many domains, a diverse network of stakeholders at different institutional levels are responsible for maintaining and evaluating shared interfaces.[31] But once the RL system comes to dominate the domain, information pruning would transform the suite of sensors used by the RL agent into the primary interface for the domain itself. This would co-opt the authority of independent agencies, and may incentivize them to redesign the domain to conform to the model specification. Mirroring

31 For an example of algorithmic models applied to child welfare risk assessment systems, see Eubanks, Virginia. *Automating Inequality: How High-Tech Tools Profile, Police, and Punish the Poor*. St. Martin's Press, 2018.

well-documented acts of malfeasance at the FAA and FCC with respect to private airlines[32] and media conglomerates,[33] this phenomenon is referred to by economists as *regulatory capture*. To avoid this outcome and respect existing norms, designers would need to document channels for data inputs whose form reflects how information is supposed to flow in particular domains. This is a difficult question, and will require independent inquiry into how RL dynamics will alter how domains work and what stakeholders may reasonably expect or want. Unless they want to make their own jobs harder, designers must find ways to translate domain-specific norms into informational guarantees and report this as part of the model specification.

## TRAINING MULTIPLE AGENTS

The application of reinforcement learning to multi-agent systems (MARL) is one of RL's most promising and rapidly growing branches. In this setup, multiple agents are modeled as acting and learning within a single environment. The agents are either learning how to achieve a single shared goal, how to behave alongside other agents with distinct goals, or how to compete for a single goal, among other scenarios. In MARL, the agents must learn what and how other agents are learning, and incorporate this knowledge into their own behavior.

### Goodhart's Law

From a design standpoint, MARL is a double-edged sword. On a technical level, it represents the culmination of efforts to simulate mechanisms for intelligent behavior that have eluded previous generations of AI research. In particular, MARL tries to generate agents that are proactively and deliberately "social." Rather than treating other agents as passive aspects of the environment, MARL agents are sensitive to each other's intentions and capabilities, and use this as grounds for actions taken. On the other hand, a strong commitment to MARL means that these social cues have been incorporated via automatic behavioral feedback, rather than subject to the designer's discretion. As such, MARL serves as RL's answer to Goodhart's Law,[34] the idea that efforts to establish control over specific metrics may backfire once other agents learn to strategically respond to the intervention. By designing agents that can anticipate each

32 Neuman, J. "FAA's 'culture of coziness' targeted in airline safety hearing." *Los Angeles Times*, April 4, 2008. URL: https://www.latimes.com/travel/la-trw-airlines4apr04-story.html

33 Travis, Hannibal. "Of blogs, ebooks, and broadband: Access to digital media as a First Amendment right." *Hofstra L. Rev.* 35 (2006): 1519.

34 The term comes from Hoskin, Keith. "The 'awful idea of accountability': inscribing people into the measurement of objects." *Accountability: Power, ethos and the technologies of managing* 265 (1996).

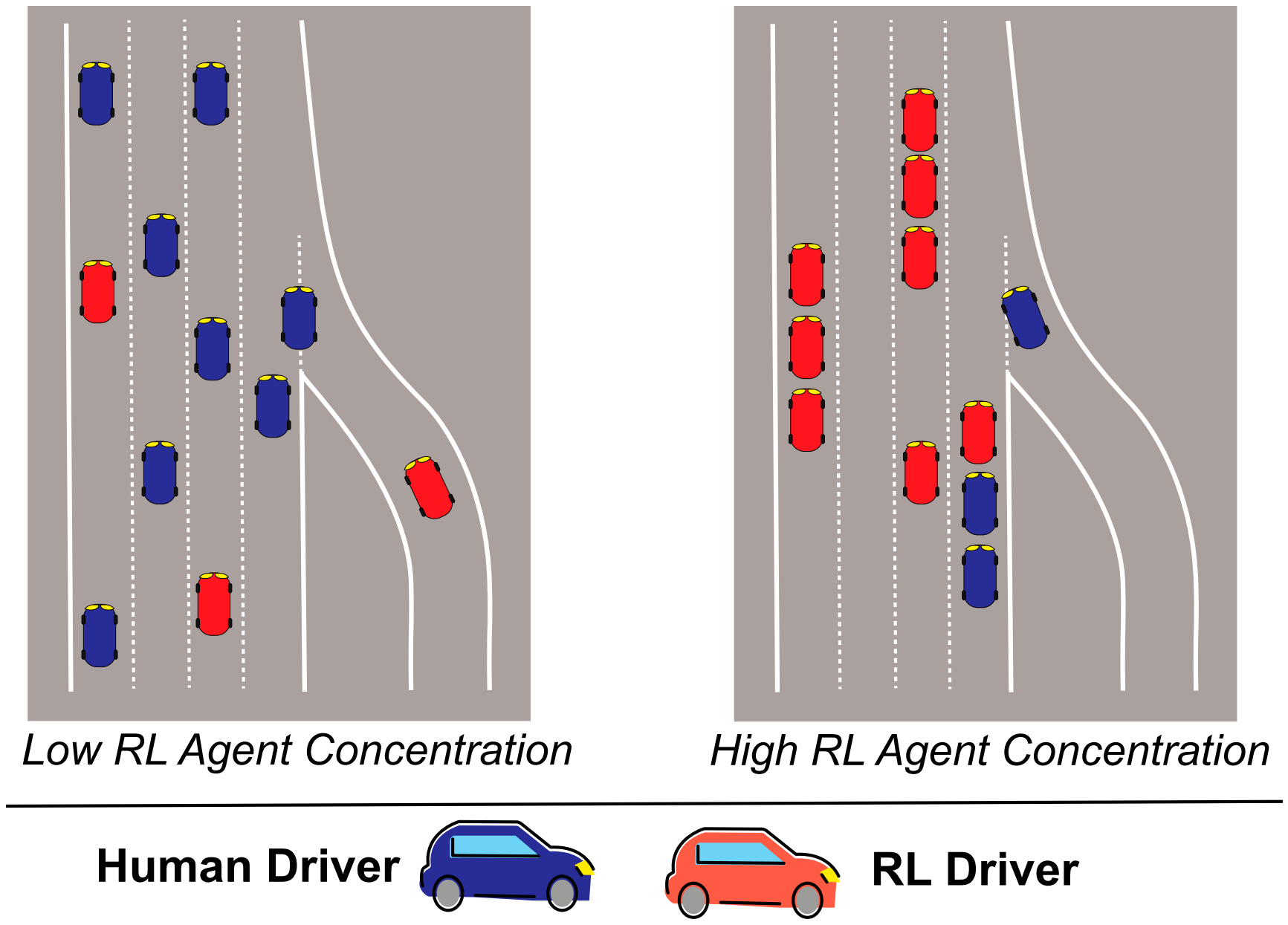

**Figure 7:** Multi-agent RL in traffic (with risk of Goodhart's Law implied). On the left, RL-based agents adopt behaviors that conform to the existing traffic flow. On the right, the learning-based agents redefine the flow of cars to optimize their own behavior and, in turn, the environment.

other's behavior, MARL tries to incorporate phenomena that would otherwise be observed as exo-feedback into the reach of behavioral feedback. A potential manifestation of this dynamic, in the case of autonomous vehicles redefining traffic flow, is illustrated in Figure 7.

Goodhart's Law is already a well-known phenomenon to machine learning researchers in applications like credit scoring, and has inspired a new branch of technical research dedicated to controlling for its effects.[35] Despite its promise, MARL may exacerbate Goodhart's Law as agents rapidly respond to each other's behavioral policies before human regulators can intervene. A harbinger of this dynamic was the "flash crash" of 2010, in which trading algorithms fed off each other's optimized policies and briefly generated a trillion-dollar market loss.[36]

To mitigate these systemic risks and make effective use of MARL's potential, three design interventions are needed. First, designers need to demonstrate the viability of a MARL setup, based on knowledge of domain-specific dynamics and how the technical results compare to

35 Perdomo, Juan, Tijana Zrnic, Celestine Mendler-Dünner, and Moritz Hardt. "Performative prediction." In *International Conference on Machine Learning*, pp. 7599–7609. PMLR, (2020).

36 Kirilenko, Andrei, Albert S. Kyle, Mehrdad Samadi, and Tugkan Tuzun. "The flash crash: High-frequency trading in an electronic market." *The Journal of Finance* 72, no. 3. (2017): 967–998.

more conventional approaches (RL or otherwise). Second, designers must show that losses have been "ring-fenced," with potential spillover effects set to remain below fundamental thresholds of contagion. This may mean safeguarding trading flows in the financial system, reducing congestion within a road network, or limiting the spread of harmful information across social media channels. Third, designers must specify the number, types, and situational range of simulated environments so that regulators can validate the learned behaviors to be deployed within industries that are critical for a given domain.

For these interventions to even be possible, designers would need to exhaustively document and justify *ex ante* the capacity for agents to dynamically incorporate other agents' policies — a task presently beyond theoretical treatments of MARL.

# Application Domains

In this section, we present three near-future applications of RL: social media content recommendation, automated vehicle fleet service, and energy grid optimization. The particularities of these cases — their present institutional makeup, past structural features, and potential effects of RL feedback once integrated — reveal the relative vulnerability of each domain to specific design risks. As each application domain requires critical infrastructure, vulnerabilities must be tracked and evaluated in order for private companies' deployment of RL to safeguard the public good. Hence, these comparative mappings illuminate possible regulatory priorities and future administrative goals for various applications of RL.

| Domain | Pre-RL challenges | Major design risk | Anticipated dynamics |
|---|---|---|---|
| Social media | Informal norms, inadequate metrics | Reward Hacking | Declining user well-being, public distrust |
| Vehicle transportation | Multimodal mobility, traffic coordination | Regulatory Capture | Forms of monopoly/ monopsony power |
| Energy infrastructure | Extreme weather, diverse energy sources, unclear market roles | Goodhart's Law | Cascading power outages, inconsistent service provision |

Table 3: Relationships between current domain challenges, RL design risks, and possible effects on dynamics.

## SOCIAL MEDIA RECOMMENDATIONS

Today, Twitter, YouTube, Facebook, and Snapchat use supervised learning to match content to users based on predictions about what they are most likely to consume. These systems have a very short horizon that limits their ability to distinguish content that happens to engage from content that users actually value.[37] Content recommendation is also highly challenging due to tightly-held but informal norms of social behavior. Nuanced and rapidly-shifting patterns

37 Milli, Smitha, Luca Belli, and Moritz Hardt. "From optimizing engagement to measuring value." In *Proceedings of the 2021 ACM Conference on Fairness, Accountability, and Transparency*, pp. 714–722. (2021).

of online interaction like "shade"[38] (public criticism) and "clapbacks"[39] (sharp responses to criticism) are normal when building an online presence, but the requisite expert awareness and know-how to manage or shape these norms are missing.

The potential of RL to learn these dynamics through behavioral feedback is intriguing. One possible solution is to control for user feedback by making the recommender horizon longer, potentially allowing the system to anticipate when users are likely to log off from a site and actively intervene. Just as trying to close a tab sometimes prompts a pop-up imploring the user not to leave, one could imagine an RL recommender that learns to recognize disengaged user behaviors, based on fine-grained metrics for how long that user "lingers" on particular content. The recommender could then suggest content that nudges the user away from those situations. As systems become more advanced, intervening on a longer horizon would be a way not just to predict user engagement based on content, but to control for styles of engaged behavior that distinct users are best suited to adopt.

However, such control feedback would blur the distinction between content that is valuable to the user and recommendations that addict the user to the algorithm. Rather than merely inflating widely shared content through network effects, as popular content is then recommended to more and more people,[40] RL may encourage the spread of manipulative, low-quality content by gaming user expectations. This would transcend the unintentional spread of misinformation (content that is engaging but misleading) and instead would constitute disinformation (targeted content that is *intended* to mislead). This indicates a dangerous propensity to reward hacking, replacing entertainment and socializing with cycles of compulsive behavior and manipulation. While these dynamics already manifest on social media platforms as exo-feedback induced through sheer scale, risks may be exacerbated when strategic decisions to create content or follow others are streamlined as part of the system specification itself.

TikTok is a harbinger of these risks. Unlike Facebook or Twitter, TikTok relies exclusively on user-generated video content, and unlike YouTube, it uses a "For You" feed that automatically

38 Isaksen, Judy L., and Nahed Eltantawy. "What happens when a celebrity feminist slings microaggressive shade?: Twitter and the pushback against neoliberal feminism." *Celebrity Studies* 12, no. 4 (2021): 549–564.

39 Jacuinde, Mireya. "Queen of the Clapback: A Framing Analysis of Alexandria Ocasio-Cortez's Use of Social Media." PhD diss., 2020.

40 See, for example, Mansoury, Masoud, Himan Abdollahpouri, Mykola Pechenizkiy, Bamshad Mobasher, and Robin Burke. "Feedback loop and bias amplification in recommender systems." In *Proceedings of the 29th ACM International Conference on Information & Knowledge Management*, pp. 2145–2148. (2020).

recommends content without the user having to "follow" anyone else.[41] This means that users interact primarily with the algorithm and secondarily with each other. The social dynamics that result are concentrated around content that is likely to rapidly increase engagement across users, instead of around content hubs with a locked-in user base. Whereas the Daily Wire repackages news stories on Facebook to appeal to conservative users,[42] TikTok generates call-to-action videos like feigning attendance at political rallies (e.g., #TulsaFlop), organizing users toward new styles of collective action.[43] It is also comparatively easier for TikTok's algorithm to reshape user behaviors, incentivizing users to follow recommended trends rather than subvert the intentions of designers in unanticipated ways.[44] To date, unlike on Twitter, TikTok users have not encouraged chatbots to adopt hate speech through targeted manipulation,[45] nor strategically undermined public polls, as in the naming of Boaty McBoatface.[46]

RL stands to exacerbate TikTok's (and future platforms') approach to content recommendation, in which users interact primarily via intrapersonal rather than interpersonal feedback.[47] Although today's systems are static and updated alongside informal human moderation, RL would dynamically update as part of the specification itself. By slotting the user into behaviors whose engagement is easier to control, rather than showing whatever content happens to have worked on others, RL could "hack" users to socialize through the platform, rather than other forms of civic participation.[48] At present, declines in mental health and public trust arise due

41 Matsakis, Louise. "TikTok Finally Explains How the 'For You' Algorithm Works," Wired, June 18, 2020. URL: https://www.wired.com/story/tiktok-finally-explains-for-you-algorithm-works/

42 Hasson, Peter J. *The Manipulators: Facebook, Google, Twitter, and Big Tech's War on Conservatives*. Simon and Schuster, 2020.

43 Bandy, Jack, and Nicholas Diakopoulos. "#TulsaFlop: A Case Study of Algorithmically-Influenced Collective Action on TikTok." *arXiv preprint arXiv:2012.07716.* (2020).

44 Anderson, Katie Elson. "Getting acquainted with social networks and apps: talking about TikTok." *Library Hi Tech News* (2021).

45 Neff, Gina, and Peter Nagy. "Automation, Algorithms, and Politics / Talking to Bots: Symbiotic Agency and the Case of Tay" *International Journal of Communication* 10 (2016): 17.

46 Lane, Anne. "Boaty McBoatface poll shows how not to do community consultation." *The Conversation* (2016).

47 Bhandari, Aparajita, and Sara Bimo. "TikTok and the "algorithmized self": a new model of online interaction." *AoIR Selected Papers of Internet Research* (2020).

48 For related work, see the following: Krueger, David, Tegan Maharaj, and Jan Leike. "Hidden Incentives for Auto-Induced Distributional Shift." *arXiv preprint arXiv:2009.09153.* (2020). Evans, Charles, and Atoosa Kasirzadeh. "User Tampering in Reinforcement Learning Recommender Systems." *arXiv preprint arXiv:2109.04083.* (2021). Carroll, Micah, Dylan Hadfield-Menell, Stuart Russell, and Anca Dragan. "Estimating and Penalizing Preference Shift in Recommender Systems." In *Fifteenth ACM Conference on Recommender Systems*, pp. 661–667. (2021).

to exo-feedback between the content optimization and users' sense of self.[49/50] This capacity to induce behaviors from sharply tuned interventions rather than platform scale is typical of RL, which is able to organize behavioral feedback according to a specified regime of control feedback.

In summary, RL-empowered recommendations would keep the user on a site by making them into something that can be engaged, not by predicting engaging content. In effect, this replaces traditional editor roles played by humans with algorithms, worsening present institutional risks. Media sources are becoming less diverse, while information channels are increasingly defined by group-specific claims to authority.[51] Journalists are routinely pressured to create and modify news content based on metrics.[52] Recommendations may encourage new users to engage with specific influencer accounts, but then make them chronically dependent on influencer marketing strategies.[53] Extrapolating these trends, there may someday be a negative affinity between RL recommendations and the fairness doctrine (or any future regulation to require broadcasters to present important issues in a manner reflecting a variety of viewpoints).

Social media is already defined by an implicit feedback chain in which people are given motivation to create and consume each other's content. What is the behavioral profile of someone capable of both creating and consuming content? What features of users are prerequisite for developing these capabilities into ingrained behaviors? Is it even acceptable to design incentives for this development to proceed? To address these challenges, reward documentation is needed that explicitly defines the metrics to answer these questions, and that identifies whether controlling for these metrics is appropriate. Absent this, the long-term effects of induced user behaviors will be much harder to manage or evaluate.

49 Frost, Rachel L., and Debra J. Rickwood. "A systematic review of the mental health outcomes associated with Facebook use." *Computers in Human Behavior* 76 (2017): 576–600.

50 Hopp, Toby, Patrick Ferrucci, Chris J. Vargo, and Jolene Fisher. "Exposure to difference on Facebook, trust, and political knowledge." *Mass Communication and Society* 23, no. 6 (2020): 779–809.

51 Van Aelst, Peter, Jesper Strömbäck, Toril Aalberg, Frank Esser, Claes De Vreese, Jörg Matthes, David Hopmann et al. "Political communication in a high-choice media environment: a challenge for democracy?." *Annals of the International Communication Association* 41, no. 1 (2017): 3–27.

52 Christin, Angèle. "Counting clicks: Quantification and variation in web journalism in the United States and France." *American Journal of Sociology* 123, no. 5 (2018): 1382–1415.

53 Woods, Steven, "#Sponsored: The Emergence of Influencer Marketing" (2016). Chancellor's Honors Program Projects. https://trace.tennessee.edu/utk_chanhonoproj/1976

## VEHICLE TRANSPORTATION

Effective management of the traffic domain depends on many key features. One is multimodal mobility, requiring a diverse set of road users (e.g., pedestrians, cyclists, cars, buses, and taxis) to move through intersections without serious risk of accidents. Another is the coordination of infrastructure across different road layers (e.g., bypasses, interchanges, stops, intersections, and crossings) to avoid debilitating slowdowns or gridlock. In combination, these features help define the flow of traffic for a given region, and their history is defined by political struggles over which modes take priority and how infrastructure may corroborate or subvert this ranking.[54]

RL can be used on both these features, through distinct styles of intervention. One method is signal control, in which traffic lights are coordinated via large-scale networks to minimize urban congestion.[55] Another approach is learning complex traffic behaviors,[56] such as automated vehicles navigating four-way stops or switching highway lanes. Part of the promise of AV fleets, which could conceivably manage thousands of cars in a given metropolitan area, is to combine these interventions in a single application. RL-tuned control over the acceleration, signaling, and steering of these fleets would be able to coordinate traffic flow across specific streets, neighborhoods, urban cores, or even entire interstate highways.[57]

These research agendas are impaired by how RL assumptions may interact with the dynamics of traffic. One problem is the unprecedented volume of data needed to learn to navigate dynamic traffic situations. This is why companies developing AVs dedicate so much attention to vehicle sensors and simulation tests of route planning.[58] Another challenge lies in achieving stability of prediction over planning horizons for driving behavior that have never existed before. The rules of traffic are both incompletely defined and subject to amendment in real time by real people faced

54 A famous example is presented in Winner, Langdon. "Do artifacts have politics?." *Daedalus* (1980): 121–136.

55 Chen, Chacha, Hua Wei, Nan Xu, Guanjie Zheng, Ming Yang, Yuanhao Xiong, Kai Xu, and Zhenhui Li. "Toward a thousand lights: Decentralized deep reinforcement learning for large-scale traffic signal control." In *Proceedings of the AAAI Conference on Artificial Intelligence*, vol. 34, no. 04, pp. 3414–3421. (2020).

56 Fisac, Jaime F., Eli Bronstein, Elis Stefansson, Dorsa Sadigh, S. Shankar Sastry, and Anca D. Dragan. "Hierarchical game-theoretic planning for autonomous vehicles." In *2019 International Conference on Robotics and Automation (ICRA)*, pp. 9590–9596. IEEE, (2019).

57 Vinitsky, Eugene, Kanaad Parvate, Aboudy Kreidieh, Cathy Wu, and Alexandre Bayen. "Lagrangian control through deep-rl: Applications to bottleneck decongestion." In *2018 21st International Conference on Intelligent Transportation Systems (ITSC)*, pp. 759–765. IEEE, (2018).

58 Sun, Pei, Henrik Kretzschmar, Xerxes Dotiwalla, Aurelien Chouard, Vijaysai Patnaik, Paul Tsui, James Guo, et al. "Scalability in perception for autonomous driving: Waymo open dataset." In *Proceedings of the IEEE/CVF Conference on Computer Vision and Pattern Recognition*, pp. 2446–2454. (2020).

with ever-evolving situations.[59] This means that the domain dynamics are significantly determined by exo-feedback, making it difficult to optimize RL traffic systems through control and behavioral feedback. Solving both problems will require designers to curate and prune information at scales that far outstrip the expertise and authority of traditional regulators.

It is worth elaborating this risk in more detail. RL behaviors occasionally exhibit *indeterminacy*, situations where the effects are surprising or difficult to evaluate.[60] As long as stakeholders can adjust to the novel behaviors or use their voice to complain about the effects, this may be manageable. But what happens when an AV fleet grows to cover an entire city? Consider the differences between a municipal entity rolling out a rideshare system, a group of startups applying for access to specific city districts, or a single giant tech company with a product that happens to be first to market. In the latter case, training agents to navigate road environments without accountability amounts to a proprietary claim on public infrastructure. The fleet operator is in essence deciding what types and magnitudes of underspecified costs are acceptable for the public to bear as the system is optimized, conducting interventions whose consequences cannot be anticipated or evaluated by critical stakeholders.

In fact, pruning information may inadvertently undercut public access to roads. AVs at a four-way stop may consider only physical distance from other vehicles as a signal, even if other information (e.g., pedestrians' gaze, drivers' hand gestures, horns honking, or common knowledge about signage) is present. But multimodal mobility depends on the ability to coordinate via different sources of information across distinct literacies (pedestrians, cars, cyclists) and sensory inputs.[61] As the AV fleets' integration grows beyond single-mode environments like highways, its interface may exclude certain road users and effectively coerce roadways to the chosen specification. As Coursera CEO and drive.ai board member Andrew Ng put it, "rather than building AI to solve the pogo stick problem [*i.e.* the presence of rare human actions], we should partner with the government to ask people to be lawful and considerate. . . . Safety isn't just about the quality of the AI technology."[62] In this case, it is not that vehicles are being automated, but that roads are being privatized.

59 This point is articulated forcefully by Pere Dubarle, as quoted in Wiener, Norbert. *The Human Use of Human Beings: Cybernetics and Society*. No. 320. Da Capo Press, 1988.

60 Heess, Nicolas, Dhruva TB, Srinivasan Sriram, Jay Lemmon, Josh Merel, Greg Wayne, Yuval Tassa et al. "Emergence of locomotion behaviours in rich environments." *arXiv preprint arXiv:1707.02286.* (2017).

61 Spickermann, Alexander, Volker Grienitz, and A. Heiko. "Heading towards a multimodal city of the future?: Multi-stakeholder scenarios for urban mobility." *Technological Forecasting and Social Change* 89 (2014): 201–221.

62 Quoted from Rodney Brooks, "Bothersome Bystanders and Self-Driving Cars". Posted July 4, 2018: https://rodneybrooks.com/bothersome-bystanders-and-self-driving-cars/

It follows that the designer must choose to either commandeer roads to conform to the pruned RL specification, or deprioritize multimodal road participants. The former amounts to a proprietary claim on public infrastructure, a form of *monopoly power*.[63] In this case, a single firm would act as the primary supplier of road services and could select a definition of optimal behavior that does not account for different interpretations of "good" driving (e.g., refrain from hitting objects, maximize fuel efficiency, or minimize travel time). Potholes are one example of this risk, as the AV fleet will add to wear and tear in specific places as certain highways are discovered to be safer or more efficient for routing purposes. In the latter case, pruning information constitutes *monopsony power*, formally defined as the exclusive "buyer" of some good or service in a particular market.[64] The service in this case is the distributed labor force (including regulators, manufacturers, and municipal bodies) compelled to support the AV firms' chosen specifications and sensory inputs. Jaywalking is a clear historical analogue, as pedestrians came to be seen as a problem for cars to avoid and largely ceded public control of streets to them.[65]

Consequently, the problem of traffic optimization using RL poses unique governance challenges beyond the hoary frame of "trolley problem" moral dilemmas. RL can assign scalar value to domain features that at present are governed in fundamentally different ways (e.g., potholes are bad but publicly managed, road rage is rude but tolerated, manslaughter is illegal). However, RL's specific role depends on the underlying philosophy that guides how the AV fleet's computation stack has been organized, for example end-to-end modeling versus more constrained approaches to "behavioral cloning" (i.e., mimicking human demonstrations[66]). Meanwhile, an AV that learns an optimally safe policy based on defensive driving may alter the flow of road intersections, destabilizing traffic even if the AV itself is physically safe. The problem is that establishing controlled interventions in one setting may generate exo-behaviors beyond what designers know how to model, unaccountably affecting the likelihood of delayed commuter times or traffic jams.

Desired features of roads, such as safety and equal accessibility, reflect that they are public and meant to be used by everyone. In addition, the overall stability of the road network weighs more heavily than the calculation of individual driver utilities as a desired metric. If RL planning

63 Lerner, Abba P. "The concept of monopoly and the measurement of monopoly power." *The Review of Economic Studies* 1, no. 3 (1934): 157–175.

64 Ashenfelter, Orley C., Henry Farber, and Michael R. Ransom. "Labor market monopsony." *Journal of Labor Economics* 28, no. 2 (2010): 203–210.

65 This story is told in Norton, Peter D. *Fighting Traffic: The Dawn of the Motor Age in the American City*. MIT Press, 2011.

66 See, for example, Bühler, Andreas, Adrien Gaidon, Andrei Cramariuc, Rares Ambrus, Guy Rosman, and Wolfram Burgard. "Driving through ghosts: Behavioral cloning with false positives." In *2020 IEEE/RSJ International Conference on Intelligent Robots and Systems (IROS)*, pp. 5431–5437. IEEE, (2020).

horizons incorporate the conditions for road stability, designers must document how these features are backed up in the chosen specification of states and actions. This is because the designer has become a steward of the road network and is now responsible for its integrity.

## ENERGY INFRASTRUCTURE

Energy infrastructure has one major advantage over other domains: the flow of electricity is determined primarily by physics. However, the problem space of energy service provision is rapidly changing, as are the tools for optimizing it. Extreme weather events related to long-term climate change have interfered with the reliable provision of service. This has led to either planned shutdowns in response to anticipated weather events (such as power utility PG&E's actions during the California wildfire season[67]) or unplanned blackouts in the aftermath of an event (such as the 2021 Texas power crisis[68]). Grids are also diversifying their sources of energy generation, often incorporating both renewables and fossil fuels to serve millions of consumers. Meanwhile, new tools for monitoring and meeting different types of demand are making it easier to observe and control energy distribution in real time for distinct consumer profiles.[69]

In principle, RL's capacity for dynamic planning seems like a natural fit for these conditions. Indeed, Google's decision to use RL to optimize a cooling system for its data centers was one of RL's first major real-world applications.[70] This is significant, as more advanced power grid optimization techniques can incorporate more behaviors of energy market participants over longer time horizons. The system could leverage additional features, such as seasonal temperature and electric vehicle charging, to reallocate energy service in real time. As long as there is a consensus definition of what the energy grid is for and whom it is supposed to serve, RL could be integrated without risk of reshaping the environment.

67 Wong-Parodi, Gabrielle. "When climate change adaptation becomes a "looming threat" to society: Exploring views and responses to California wildfires and public safety power shutoffs." *Energy Research & Social Science* 70 (2020): 101757.
68 Kemabonta, Tam. "Grid Resilience analysis and planning of electric power systems: The case of the 2021 Texas electricity crises caused by winter storm Uri (#TexasFreeze)." *The Electricity Journal* 34, no. 10 (2021): 107044.
69 Ardakanian, Omid, Vincent WS Wong, Roel Dobbe, Steven H. Low, Alexandra von Meier, Claire J. Tomlin, and Ye Yuan. "On identification of distribution grids." *IEEE Transactions on Control of Network Systems* 6, no. 3 (2019): 950–960.
70 Knight, Will. "Google just gave control over data center cooling to an AI." "MIT Technology Review, August 17, 2018. URL: https://www.technologyreview.com/2018/08/17/140987/google-just-gave-control-over-data-center-cooling-to-an-ai/

However, Google's test application still depended on two-layer verification based on safety constraints defined by human operators.[71] Without such oversight, the behavioral feedback unlocked by RL may generate inappropriate interventions. Imagine systems that automatically shut off consumer power during high load periods, or pricing shifts that happen every hour. Just as the Texas energy crisis generated exorbitant charges due to critical supply shortages, RL may enable control paradigms whose assumptions are grossly mismatched with the diverse behaviors of consumers, utilities, regulators, and renewable power generators. As such, the optimization of energy provision has a powerful affinity with Goodhart's Law.

Consider the case of the South Australian blackout of 2016. Extreme weather combined with reduced wind farm output to significantly reduce the capacity of the South Australian distribution grid. In response, the system automatically tried to redirect power from Victoria to South Australia in order to meet demand. This in turn overloaded the Heywood Interconnector between the two states, isolating the South Australian power system and leading to a so-called Black System Event: the total failure of the distribution grid.[72] At the heart of this complex energy crisis was the unexpected behavior of certain statewide energy market participants relative to the assumptions of the system designers. More specifically, the compliance of key energy stakeholders with the National Electricity Rules was ambiguous with respect to abnormal conditions. This led to a series of critical communication failures between market operators and generators that caused a catastrophic (rather than controlled) failure of the grid.[73]

These types of scenarios are likely to become more common as grid optimization techniques improve. As more stakeholder behaviors are brought into scope, the potential for critical failure increases due to the risk of failing to fully incorporate the exo-feedback between stakeholder roles. RL has a significant bias toward regular and predictable dynamics, and energy optimization promises to play into control and behavioral feedback through rich amounts of data whose covariates and features are easy to catalogue and trace. However, it is unclear whether RL techniques can be made resilient in the face of unexpected weather events, and how different approaches to optimization might affect strategic interactions between residents, regulators, and other constituents over time (let alone during states of emergency).

71 Gasparik, A., C. Gamble, and J. Gao. "Safety-first ai for autonomous data centre cooling and industrial control." *DeepMind blog* August 17, 2018. URL: https://deepmind.com/blog/article/safety-first-ai-autonomous-data-centre-cooling-and-industrial-control

72 Yan, Ruifeng, Tapan Kumar Saha, Feifei Bai, and Huajie Gu. "The anatomy of the 2016 South Australia blackout: a catastrophic event in a high renewable network." *IEEE Transactions on Power Systems* 33, no. 5 (2018): 5374–5388.

73 The Black System Event Compliance Report. Available online: https://www.aer.gov.au/system/files/Black%20System%20Event%20Compliance%20Report%20-%20Investigation%20into%20the%20Pre-event%20System%20Restoration%20and%20Market%20Suspension%20aspects%20surrounding%20the%2028%20September%202016%20event.pdf

As a result, RL is likely to introduce new political dynamics to energy grids. On the one hand, RL more easily provides marginal cost benefits than service resiliency, making its consumer benefits more obvious than its public ones. This is far removed from the Rural Electrification Act of 1936, which funded cooperative electric power companies to provide electricity to poor areas previously thought to be unserviceable.[74] On the other hand, longer planning horizons may permit regulatory bodies to make new decisions about their own energy policy, in tandem with economic or environmental commitments. Cities could also decide to redistribute efficiency gains by improving street lighting or providing preferred energy prices to low-income residents.

Because RL could be used to throttle energy services just as easily as provision them more equitably, its technical proficiency should not be conflated with social desirability. Rather, RL will bring public policy commitments more sharply into focus and allow constituents to make more strategic decisions about energy generation and consumption. Either way, the responsibility of RL designers will be to catalogue alternative specifications and communicate to energy regulators how choices about optimization are likely to reshape service provision over time.

74 Cooke, Morris Llewellyn. "The Early Days of the Rural Electrification Idea: 1914–1936." *American Political Science Review* 42, no. 3 (1948): 431–447.

# Reward Reports

Evaluating which reward structures could be institutionally affirmed and technically supported through feedback requires asking what it would mean to govern RL systems, not just optimize them. At present, much of the appeal of RL is that practitioners can use many types of tools to design reward functions (such as scaling, adding terms, changing the meaning of features, or behavior cloning). But this variety of tools has outpaced our understanding of their distinctive effects. Because mechanisms to audit RL systems pre- and post-deployment are missing, designers are unable to map how one type of reward function (and other salient aspects of RL specification) will interact with a given domain. Reward functions can also be expressed in multiple ways at different scales, and it is unrealistic to hold practitioners responsible for articulating all possible representations from particular design decisions independently.

In theory, RL permits designers to "simulate" public policy by either adding features or maximizing expected utility at arbitrary computation scales. Such designers are hypothetically in a better position to define and measure value than either the market or political institutions. Rather than allowing normative concerns to be expressed "suboptimally" by boycotting a service or making new zoning laws, it would simply be more efficient to let AV designers "figure out" what the reward function is for driving in a given city through a mix of routing adjustments and dynamic pricing. However, these assumptions are flawed, as a complete catalogue of relevant features beyond the physical road environment is not available for AV firms to optimize. If this form of computational governance were seriously pursued, it would eclipse a core function of political governance and democratic dissent: articulating underspecified normative criteria that distinguish between utility losses and the definition of good social outcomes. This is where questions of reward optimization meet the wider themes of AI governance.

The mapping from function to domain will require more detailed knowledge about particular use cases, the range of alternative specifications that were technically feasible prior to deployment, and the types of feedback available to help optimize the system following deployment. In this section, we propose *Reward Reports* as a form of *ex ante* documentation supporting these requirements, empowering stakeholders to deliberate about how RL specifications and types of feedback are to be structured to work in support of human ends.

Building on proposals for "model cards" for deployed machine learning models,[75] as well as "datasheets for datasets" to document the use of data in particular high-stakes domains,[76] Reward Reports will detail intended and demonstrable performance characteristics of trained and untrained RL specifications within particular domains. These points are outlined in Table 4.

| Name | Pertains to | Development phase | Key features |
|---|---|---|---|
| Model card | Learned model | ex post | recommended use, core metrics, known ethics issues |
| Datasheet | Data structure | ex post | motivation, composition, collection process, uses |
| Reward report | Optimization problem | ex ante and ex post | domain assumptions, horizon, specification components |

Table 4: Modes of documentation for AI systems, including development phase and key features.

Reward Reports draw attention to the risks entailed in defining what is being optimized, which is especially consequential for RL-formulated problems. Documentation of states, actions, and rewards as they bear on the lifestyles and opportunities of human beings — whether end-of-life healthcare, traffic accident avoidance, or some other context — should substantively inform whether explicit regulation is needed and what form is warranted for the specific risks at stake. Making sense of this documentation will entail renewed interpretation and application of particular legal concepts so that new types of RL feedback corroborate the purpose and integrity of human domains. For these reasons, Reward Reports, based on datasheets and model cards from related domains, could serve as a third, complementary form of documentation, giving both designers and regulators the tools needed to evaluate a given AI system.

The problem of defining what is being optimized is implicit in any ML system. The three basic elements of any machine learning task are the data, model, and optimization problem (i.e. reward). Incorporating a given optimization, achieved automatically in RL via behavioral feedback, is often done manually by human engineers as part of a static prediction task in supervised or unsupervised learning. In effect, RL's automation of this system component outsources the human evaluation to institutional layers beyond the scope of the task, falling

75 Mitchell, Margaret, Simone Wu, Andrew Zaldivar, Parker Barnes, Lucy Vasserman, Ben Hutchinson, Elena Spitzer, Inioluwa Deborah Raji, and Timnit Gebru. "Model cards for model reporting." In *Proceedings of the conference on fairness, accountability, and transparency*, pp. 220–229. (2019).

76 Gebru, Timnit, Jamie Morgenstern, Briana Vecchione, Jennifer Wortman Vaughan, Hanna Wallach, Hal Daumé III, and Kate Crawford. "Datasheets for datasets." *arXiv preprint arXiv:1803.09010.* (2018).

under the purview of system managers, law courts, or regulatory bodies. It is these entities that must decide whether or how the optimized behaviors align with the application domain, in correspondence with resultant risks and possible harms. Just as RL's unique design features force designers to be explicit up front about the terms of the chosen optimization, it also highlights the need for these institutions to take up aspects of AI governance more broadly via corresponding legal regimes and organizational priorities. While Reward Reports are needed to determine how the emergent behavior of RL systems tracks with the intent of designers, we also present them as a structured curriculum for these institutions to learn how to oversee ML systems more broadly.

## THE CORE COMPONENTS OF REWARD REPORTS

Our presentation of Reward Reports draws inspiration from recent work for supervised machine learning that focuses on model transparency and impact assessments,[77] as well as documentation of performance across deployment conditions, domains, and specified features.[78] However, our concerns move beyond learned models toward the emergent behaviors that are adopted by the RL agent as it learns to navigate its environment, and that are responsible for the risks we have catalogued. There are three primary reasons for this shift in concern.

First, designers need a regular "change log" that tracks the curriculum by which RL agents have adopted specific behaviors, i.e., a mechanism for revisiting the specification if the system is found to "hack" the domain in unanticipated ways. Documentation of learned models assumes that accuracy thresholds on salient features serve as a sufficient proxy for domain suitability. A core theme of this whitepaper has been that static thresholds are insufficient to capture the real-world performance of RL systems, which are capable of learning dynamics whose evaluation requires more exhaustive documentation. Designers also need to specify other curricula, like excluded features or reward metrics, that were considered, but ultimately rejected as unsuited to the domain in question. This would counteract the system's propensity to double-down on features the designer prioritized over other alternatives. Furthermore, Reward Reports could be consulted and updated periodically as

77 Reisman, Dillon, Jason Schultz, Kate Crawford, and Meredith Whittaker. "Algorithmic impact assessments: A practical framework for public agency accountability." *AI Now Institute* (2018): 1–22.
78 Mitchell, Margaret, Simone Wu, Andrew Zaldivar, Parker Barnes, Lucy Vasserman, Ben Hutchinson, Elena Spitzer, Inioluwa Deborah Raji, and Timnit Gebru. "Model cards for model reporting." In *Proceedings of the conference on fairness, accountability, and transparency*, pp. 220–229. (2019).

part of ongoing reviews, with the frequency depending on particular domain features and vulnerabilities.

Second, Reward Reports would provide support to at-risk populations for litigation via corresponding branches of law, comprising a community feedback channel for design decisions. For example, in the case of liability, some scholars have argued that the complex relationship between design intent, ML system opacity, and administrative negligence prohibits a clear application of legal criteria for harms.[79] Others claim that liability law is well equipped to deal with these issues,[80] while other opinions are more mixed in their assessment.[81] RL is likely to increase these ambiguities by automating different types of feedback and blurring important boundaries between design choices and emergent system behaviors. Our vision of Reward Reports as “design receipts” reduces this ambiguity by allowing exo-feedback to flow directly between impacted communities and design choices. We expect Reward Reports would also fill an important gap in the ability of policy-adjacent organizations to understand what RL systems are capable of, and could be used to inform new standards for performance evaluation.

Third, Reward Reports are needed for policymakers to come to a decision about whether the system’s emergent behaviors align with how domains are supposed to work. Model cards focus on a modular piece of AI system development (e.g., face detection) that can be abstracted from particular use cases. Because of this, they are primarily useful to fellow ML engineers and AI designers, who can rely on prior familiarity with the use context to evaluate whether the model is appropriate or not. But RL by definition encompasses a larger system within itself, sensitive to multiple forms of feedback. RL “cards” would therefore apply to each specific deployment, rather than function as something modular that could be confidently applied elsewhere with little loss of meaning. As we have argued, the risks of specific RL deployments are high enough — and shifts in response to feedback sufficiently likely — that explicit expert judgment about how specific use cases may fit into and transform domain dynamics is paramount.

A more complete taxonomy of Reward Reports is presented below:

79 See the following literature: Scherer, MU, “Regulating Artificial Intelligence Systems: Risks, Challenges, Competencies, and Strategies” (2015) 29 Harv. JL & Tech. 353.

80 Cauffman C, “Robo-Liability: The European Union in Search of the Best Way to Deal with Liability for Damage Caused by Artificial Intelligence” (2018) 25 Maastricht Journal of European and Comparative Law.

81 Morgan J, “Torts and Technology” in Roger Brownsword, Eloise Scotford, and Karen Yeung (eds), *The Oxford Handbook of Law, Regulation and Technology* (2017).

**REWARD REPORT CONTENTS**

**System Details:** Basic system information.

- Person or organization developing the system
- Deployment dates
- Contact

**Optimization Intent:** The goals of the system and how reinforcement manifests.

- Goal of reinforcement
- Performance metrics
- Oversight metrics
- Failure modes

**Institutional Interface:** The interconnections of the automated system with society.

- Involved agencies
- Stakeholders
- Computation footprint
- Explainability
- Recourse

**Implementation:** The low-level engineering details of the ML system.

- Reward, algorithmic, and environment details
- Measurement details
- Data flow
- Limitations
- Engineering artifacts

**Evaluation:** Specific audits on system performance.

- Evaluation environment
- Offline evaluations
- Evaluation validity
- Performance standards

**System Maintenance:** Plans for long-term verification of behavior.

- Reporting cadence
- Update triggers
- Changelog

As argued below, the regulatory challenges associated with control, behavioral, and exo-feedback are matched by particular branches of law. Each of these has a well-developed set of tools, tests, and standards that could evaluate a given reward function with reference to the domain in which it will operate. Reward Reports would make this evaluation possible by documenting how design choices relate to the anticipated dynamics. Such documentation would clarify regimes of control and responsibility for AI firms, courts, and regulators in instances of litigation and arbitration. These points are analytically summarized in Table 5.

| Type of feedback | Associated risk | Legal criterion | Example standards |
|---|---|---|---|
| Control | Regulatory Capture | Antitrust | Rule of reason, infrastructural regulation |
| Behavioral | Reward Hacking | Liability | Foreseeability, product liability |
| Exogenous | Goodhart's Law | Administrative | Interoperability, state action doctrine |

**Table 5:** Relationships between types of feedback, branches of law, and exemplary standards.

## ANTITRUST

The goal of antitrust law is to prevent firms from becoming powerful enough to enact barriers that interfere with market competition. These barriers may take various forms. Dominant firms may gouge prices to take advantage of consumers' access to necessities. They could also delimit potential rivals' capacity to enter the market through private cartels or standing agreements with captured regulatory interests. Firms can also merge with each other to reduce (or outsource) production costs. It is now also possible to "platform" specific services to lock in users and extort other companies for access.[82] These cases are defined by insufficient checks and balances in the market, in essence allowing the firm to act as a tyrannical form of private government. As such, the underlying purpose of antitrust is not to "keep markets free," but to mitigate the social vulnerabilities that result from arbitrary concentrations of private power.[83]

Antitrust has a natural affinity with the monitoring of control feedback. Just as the RL designer is concerned with the agent's ability to choose effective actions based on observed states, antitrust enforcement is concerned with firms' ability to manipulate market dynamics. The danger manifests when control feedback, for the purpose of mastering task completion, comes to serve as a proxy for control of critical social domains. Consider a self-driving car firm that wants to use RL not just to manage local fleet behavior, but to optimize traffic across an entire city grid. In this case, there is a need to match regimes of control to forms of social welfare such as low prices, acceptable rates of congestion, or demonstrably equitable platform access.

How might this standard be applied to automated feedback systems like RL? The intellectual history of antitrust is marked by distinctive tests for market power. For several decades, antitrust law in the United States has been dominated by the consumer welfare standard: the view that corporate mergers are not harmful to consumers unless the quality of goods declines or prices increase.[84] The so-called "New Brandeis movement" argues that this standard has generated a myopic understanding of structural harms and ignores the nuanced strategies adopted by firms to skirt regulatory tests.[85] There is active legal and economic discussion on which tests for market power are most rigorous, the conditions under which quantitative thresholds of harm are appropriate, and what kinds of expert testimony should

82 Morton, Fiona M. Scott, and David C. Dinielli. *Roadmap For An Antitrust Case Against Facebook*. Technical report, Omidyar Network, June 2020.

83 Khan, Lina. "The New Brandeis Movement: America's Antimonopoly Debate." *Journal of European Competition Law & Practice* 9, no. 3 (2018): 131–132.

84 Kovacic, William E. "The Antitrust Paradox Revisited: Robert Bork and the Transformation of Modern Antitrust Policy." *Wayne L. Rev.* 36 (1989): 1413.

85 Khan, Lina M. "Amazon's Antitrust Paradox." *Yale lJ* 126 (2016): 710.

decide these tests' applicability to particular cases.[86] We anticipate that the capacity of RL to leverage different types of feedback automatically will make antitrust enforcement particularly challenging, as systems may learn to commandeer the domain in ways the firm itself does not understand, let alone regulators.

Reward Reports will help make this enforcement possible by empowering legal persons to apply the *rule of reason* to specifications that inappropriately restrict competition in the domain. A complete Reward Report would document counterfactual designs of states, actions, and rewards for alternative specifications, in order to provide justification for revisiting and altering the chosen specification post-deployment. Regulatory bodies would then be able to monitor problematic types of feedback and match relevant antitrust standards to corresponding technical decisions and data pipelines. Reward Reports' pragmatic style of documentation matches the substantive focus of antitrust on predatory behavior, even if applied to RL systems rather than large corporations. As was recently argued by K. Sabeel Rahman, traditional antitrust approaches were not intended to reify a particular regulatory blueprint for specific goods or services. Rather, they helped comprise a method of inquiry rooted in the ability to trace social vulnerabilities back to the behavior of powerful private agents.[87]

Once Reward Reports are widely available, several antitrust tests could be applied. Historically, antitrust has interpreted problems like reward misspecification through the *common carriage* standard.[88] For example, beyond some geofencing threshold, an AV fleet could be interpreted as a common carrier that is responsible for ensuring fair and equal access to its platform. Control feedback could then be regulated under common carrier doctrine, with performance parameters to be set or overseen by a third party. This oversight would require documentation of how designers have specified social welfare in terms of what the fleet has learned (or could learn) to optimize. There is also the standard of *infrastructural regulation*:[89] some kind of firewall or public interface could be created with the firms that produce the AV specification to ensure that it remains inclusive of road users. This regulation would prevent the fusion

86 See, for example, William Novak, Chapter 4: The Public Utility Idea and the Origins of Modern Business Regulation, in *The Corporation and American Democracy* 139 (N. R. Lamoreaux & W. J. Novak eds., Harvard University Press 2017). David Autor et al., The Fall of the Labor Share and the Rise of Superstar Firms, 135 *Q. J. Econ*. 645 (2020). Anna Stansbury & Lawrence H. Summers, Declining Worker Power and American Economic Performance, *Brookings Papers On Econ*. Activity (2020).

87 Rahman, K. Sabeel. "Infrastructural Regulation and the New Utilities." *Yale J. on Reg.* 35 (2018): 911.

88 Noam, Eli M. "Beyond liberalization II: The impending doom of common carriage." *Telecommunications Policy* 18, no. 6 (1994): 435–452.

89 Rahman, K. Sabeel. "Regulating informational infrastructure: Internet platforms as the new public utilities." *Georgetown Law and Technology Review* 2 (2018): 2.

of private service provision with roadway access via restricted information channels, while permitting external regulators to investigate sensory inputs and confirm they do not exclude mobility participants. This may become more important as we transition to "smart" roadway infrastructure that will make citywide traffic optimization more viable, as traffic control must remain public as well as optimal. These capabilities may require open APIs and standards for access that are externally enforced. Finally, Reward Reports would help clarify the application of *computational antitrust* to safety-critical domains. This emerging set of tools aims to automate and refine antitrust tests by mechanizing legal analysis, perhaps through the use of RL itself, to simulate the merger of firms.[90] AI firms and antitrust agencies would then share many techniques for system evaluation, perhaps improving their ability to keep pace with new forms of control feedback.

## LIABILITY

Liability elucidates the legal obligations of a party in terms of its responsibilities to some other party. Many liability regimes exist, ranging from civil law (including torts), to criminal law (including various types of punishment). Liability interprets harms or damages conducted in the context of binding agreements that stipulate how they can be rectified or paid back. This allows cases to be resolved and specifies litigation costs that may help incentivize powerful firms to avoid harmful behavior in the first place.[91]

For RL systems, liability claims will hinge on the types of behavioral feedback incorporated. As an agent learns to navigate its specified environment, it may adopt behaviors that are optimal yet harmful to stakeholders — for example, the interests of marginal subpopulations left out of the design scope but implicated in the system behavior. Different interpretations of product liability may be relevant here. According to *strict liability*, producers of a given RL system would be responsible for any harm it generates even in the absence of designer intent. But according to *negligence liability*, designers could be found at fault only if they were found to have acted irresponsibly in light of foreseeable and preventable harms.[92] Consequently, if a claimant wants to file suit against a given firm (e.g., for reducing neighborhood housing prices due to the

90 Schrepel, Thibault, "Computational Antitrust: An Introduction and Research Agenda" (January 15, 2021). Stanford Computational Antitrust (Vol. 1) 2021, Available at SSRN: https://ssrn.com/abstract=3766960

91 Polinsky, A. Mitchell, and Steven Shavell. "Legal error, litigation, and the incentive to obey the law." *JL Econ. & Org.* 5 (1989): 99.

92 Zech, Herbert. "Liability for AI: public policy considerations." In *ERA Forum*, vol. 22, no. 1, pp. 147–158. Springer Berlin Heidelberg, 2021.

dynamic behavior of AV fleets), there would have to be a way to tie the RL behaviors to specific liability regimes that conform to the standards of care established for that application domain.

Establishing negligence has proven difficult in the context of AI systems, let alone RL behaviors. Consider the now-familiar notion of "black box" models generated via deep learning. Classifiers with hidden layers or many parameters often generate decisions whose criteria cannot be traced to a particular point of model training. As a result, their predictions cannot be fully accounted for or explained, making them difficult to audit and requiring a new basis on which to evaluate models, data, vendors, or administrators as liable for any resulting forms of harm.[93] Another issue is reconciling distinct theories of product liability in the context of particular AI applications. Beyond negligence, these include design defects, manufacturing effects, failure to warn, misrepresentation, and breach of warranty, any of which may apply to the harms generated by an AI system, depending on how its model was trained or its data procured.[94]

By default, RL is likely to further muddle these distinctions due to the dynamics its behavior introduces into the specification, or generates outside it. Even if some blame falls on the classifier's constraints, design choices, or available data, RL algorithms also iteratively update behaviors over time based on multiple types of feedback. This may destabilize the relationship between manufacturer, consumer, and product, possibly creating gaps in liability for RL-generated social harms.[95] Beyond the exo-feedback often at stake in black box algorithms, more specific standards of care are needed for RL behavioral dynamics.

Reward Reports will concretely address these challenges in two ways. First, they will make risky interactions between learned behaviors and domain dynamics more *foreseeable*, permitting new standards of care that will allow design decisions to be interpretable as negligent. Causation tests for automated systems often fail because they depend on foreseeability that black box models render moot. For example, tests for proximate cause address negligence, but depend on inquiry into what AI creators could have demonstrably anticipated about the system's performance. Meanwhile, other tests focus on a causation threshold that would tie an AI's decisions to concrete harms, though this requires thorough documentation of the AI's reasoning as demonstrably faulty. In both cases, absence of foreseeability makes it impossible to assign responsibility for

93 Yu, Ronald, and Gabriele Spina Alì. "What's inside the Black Box? AI Challenges for Lawyers and Researchers." *Legal Information Management* 19, no. 1 (2019): 2–13.
94 Villasenor, John. "Products liability law as a way to address AI harms." Brookings Institution, October 31, 2019. URL: https://www.brookings.edu/research/products-liability-law-as-a-way-to-address-ai-harms/
95 Swanson, Greg. "Non-Autonomous Artificial Intelligence Programs and Products Liability: How New AI Products Challenge Existing Liability Models and Pose New Financial Burdens." *Seattle UL Rev.* 42 (2018): 1201.

system behavior, even where harms are in principle traceable to design choices. But greater transparency could shift the burden of liability onto the producer of a given system if the reasoning behind these choices is made explicit, permitting a "sliding scale" of foreseeability.[96]

By supplying *ex ante* documentation, Reward Reports render resultant harms more foreseeable and the agents' learned behaviors more explicable in light of the specification. In fact, Reward Reports could include a "sliding scale" of counterfactual specifications that analogize between reward function and salient domain risks. As designers must decide what the agent is trying to optimize, as well as the state-action space it is learning to navigate, care must be taken to track how alternative specifications map onto domain risks so that design decisions lie beneath the threshold for negligence. That threshold itself is dependent on domain risks, and may be lower (i.e., easier to meet the burden of proof for a claim) in safety-critical settings where resultant harms are more easily traceable to design choices.

Second, Reward Reports will help *reconcile* the application of distinct liability standards to system behaviors by clarifying which stakeholders shoulder the burden for harms resulting from behavioral or exo-feedback. Consider the case of an AV fleet that has learned to game the traffic signals of a residential neighborhood, disrupting traffic in ways that other road users must accommodate. Is this the fault of designers for failing to document this behavior in consequent Reward Reports? Or the fault of the city planners who allowed the fleet to be deployed, despite advance documentation of risks? Or the fault of fleet managers who chose to hide this documentation from other stakeholders as the behavior became foreseeable? Rather than different standards for negligence, these sorts of questions bear on distinct interpretations of the system behavior as inherently defective, intentionally misrepresented, or the result of a communication failure between administrators.[97]

Reward Reports will address this indeterminacy by documenting how sources of feedback have been incorporated into learned behaviors. While possible behaviors should be fully catalogued prior to deployment, foreseeability of outcomes also depends on closely monitoring and accounting for emergent behaviors as the agent learns to navigate its specified environment. In particular, reports must distinguish which elements of the specification from which the agent has automatically learned a policy have been reported to or rejected by those subject to its post-deployment effects. In this way, successive reports will clarify who is responsible for which elements of the specification and the reasons for updating them. It follows that this

96 Bathaee, Yavar. "The artificial intelligence black box and the failure of intent and causation." *Harv. JL & Tech.* 31 (2017): 889.
97 Villasenor, John. "Products liability and driverless cars: Issues and guiding principles for legislation." (2014).

documentation must also include a commitment to periodically monitor performance post-deployment, as controls and behavior respond to and interact with exo-feedback.

## ADMINISTRATIVE LAW

Executive agencies are tasked with enforcing the laws passed by legislative bodies. The various emergent issues these agencies face related to rulemaking and standard-setting are addressed by administrative law, which regulates the relationships between agency functions and legal persons beyond what statutes make explicit. In the words of prominent legal scholar Jennifer Cobbe, the guiding principle of administrative law is that "the more serious and consequential a decision the greater the need to give reasons,"[98] meaning that a critical function of state agencies is to continuously monitor and justify the law's applicability to domains under their purview.

Administrative law can be thought of as the law's answer to exo-feedback. An RL-enabled system may drift the application domain over time due to dynamics outside its specification; likewise, a federal executive agency is tasked with protection and regulation of a public domain beyond what statutes are able to specify. There is also a risk that private firms, in their use of AI systems, cause domains to drift in ways that undercut this critical agency function. For example, if a company shows that traffic lights can be readily optimized using RL, it could ask for more direct control of public infrastructure, but this may well change the domain. These cases must be arbitrated so that the respective roles of public agencies and private firms are appropriately distinguished. Per administrative law, the firm in question may have to perform the duties of a public regulator, such as fulfilling Freedom of Information Act requests.[99]

There is thus a need for judicial review of automated decision-making: identifying the conditions under which a private entity may manage ML (or RL) systems without undercutting the authority of public agencies. However, the private use of AI systems is already permitted for many domains, in effect automating the roles of critical domain personnel. Algorithms now help manage Medicaid records, unemployment services, and credit score allocation, among many other administrative services.[100] In these cases, private companies act as AI service

98 Cobbe, Jennifer. "Administrative law and the machines of government: judicial review of automated public-sector decision-making." *Legal Studies* 39.4 (2019): 636–655.

99 Van Loo, Rory. "The New Gatekeepers: Private Firms as Public Enforcers." *Va. L. Rev.* 106 (2020): 467.

100 For examples, see Pasquale, Frank. "Data-informed duties in AI development." *Colum. L. Rev.* 119 (2019): 1917. Amine, Samir, and Pedro Lages Dos Santos. "Technological choices and unemployment benefits in a matching model with heterogenous workers." *Journal of Economics* 101, no. 1 (2010): 1-19. Wang, Gang, Jinxing Hao, Jian Ma, and Hongbing Jiang. "A comparative assessment of ensemble learning for credit scoring." *Expert systems with applications* 38, no. 1 (2011): 223–230.

vendors that provide critical data analytics for the purpose of government procurement. However, the executive agency may have outsourced some of its authority to the vendor in ways that could leave exo-feedback unchecked. In RL's case, there is an additional danger that some discretionary authority has been delegated to the AI system itself, beyond the capacity of either the executive agency or private vendor to review.

While a complete list of administrative law's potential applications is beyond the scope of this paper, here we sketch the general case of government entities permitting RL-automated decision procedures in ways that are problematic. Imagine the use of an RL system that recommended the optimal policy for closing or re-opening private businesses to minimize COVID infections or related deaths.[101] Here, both state actors and private vendors are functionally ignorant of system effects — the former unaware of technical choices underlying the tool's optimization, the latter lacking knowledge of the expert judgment needed to adjudicate use cases. RL may exacerbate this administrative gap, as private firms and startups develop AI systems that automate or appropriate functions of the state without anyone knowing until it is too late.

Reward Reports will help mitigate this risk by documenting whether and in what ways designers appropriated existing public infrastructure or received direction from public officials when specifying particular systems. This will improve RL systems' *interoperability* — the coordination of information systems within and across organizational boundaries — and help lessen the burden for state agencies.[102] More specifically, Reward Reports will clarify the constituent public and private roles at stake in vendoring state functions, and the terms under which these functions are handed off to RL systems. This is particularly important in cases where an existing vendored AI system later incorporates RL in its functioning. Reward Reports would assist in assuring that this system has been periodically revisited and evaluated to conform to the originally specified administrative mandates, and has not "drifted" the use context in directions that permit easier optimization at the expense of protected rights and freedoms. Hence, Reward Reports will aid post-deployment investigations of the system by disentangling the constituent roles of vendors and state actors as the domain inevitably drifts through exo-feedback.

101 Elias, Blake, Alexander F. Siegenfeld, and Yaneer Bar-Yam. "Pandemic Response as Reinforcement Learning." NeurIPS 2020 Workshop on Machine Learning for Economic Policy. URL: http://blakeelias.name/papers/MLEconPolicy20_paper_29.pdf.
102 Matney, Susan A., Bret Heale, Steve Hasley, Emily Decker, Brittni Frederiksen, Nathan Davis, Patrick Langford, Nadia Ramey, and Stanley M. Huff. "Lessons learned in creating interoperable fast healthcare interoperability resources profiles for large-scale public health programs." *Applied Clinical Informatics* 10, no. 01 (2019): 087–095.

In the words of AI and legal scholars Kate Crawford and Jason Schultz: "The key question . . . is whether AI vendors—and the systems they create—are merely tools that government employees use to perform state functions, or whether the vendor systems perform the functions themselves."[103] In such cases, Crawford and Schultz have proposed application of the *state action doctrine*,[104] which delimits constitutional provisions to government entities, to AI systems themselves. This would enable the reinterpretation of AI systems as acting on behalf of the government, even if they are privately built and managed. In other words, systems that perform state functions would fall within courts' jurisdiction of the state action doctrine, and be prohibited from violating rights and freedoms that those agencies are also forbidden to violate. Crawford and Schultz further propose three tests for courts to reinterpret the state action doctrine in the context of state-procured AI systems: (1) the public function test, which asks whether the private entity performed a function traditionally and exclusively performed by government; (2) the compulsion test, which asks whether the state significantly encouraged or exercised coercive power over the private entity's actions; and (3) the joint participation test, which asks whether the role of private actors was "pervasively entwined" with public institutions and officials.

Reward Reports would also make design considerations more transparent to civil society groups, supplementing the administrative capacity of private firms that are considering whether to vendor an RL system. This will bolster the application of administrative law by allowing firms to more easily perform public regulator duties. Prior capacity is not necessarily needed, as NGOs, standard-setting bodies, and academic researchers could contribute to Reward Reports and thereby serve as checks on disparate impacts and other problematic outcomes.[105] Thanks to this input, firms could offer RL-enabled services without risking unaccountable exo-feedback, preserving the intent of administrative law to provide adequate reasoning for consequential decisions.

103 Crawford, Kate, and Jason Schultz. "AI systems as state actors." *Columbia Law Review* 119, no. 7 (2019): 1941–1972.
104 Not to be confused with the technical specification of state-action spaces discussed elsewhere in this whitepaper.
105 For an example, see Cath, Corinne. "Governing artificial intelligence: ethical, legal and technical opportunities and challenges." (2018): Phil. Trans. R. Soc. A 376: 20180080. http://dx.doi.org/10.1098/rsta.2018.0080

# Conclusion

The long-term goal of Reward Report documentation is domain certifications for the deployment and operation of RL systems. This will likely require Reward Reports to be standardized as part of a larger certification procedure. A preliminary step is the integration of Reward Reports to existing technical conformity assessments conducted by state agencies, and trustworthy AI frameworks developed in the private sector. For example, the AI Risk Management Framework recently developed by the National Institute of Standards and Technology[106] could be updated to reference outstanding Reward Reports of deployed systems. Over time, the domain-specific and cross-referential nature of these reports would comprise a paper trail permitting additional agencies to provide critical input and oversight over AI systems from whose design choices they are presently excluded. In particular, flowing from our discussion of antitrust, liability, and administrative law, we envision three key policy audiences for Reward Reports:

**1) Trade and commerce regulators, such as the U.S. Federal Trade Commission representatives and staffers.** As recently expressed by Commission Chair Lina Khan, FTC regulators should favor a "holistic approach to identifying harms" and "targeting root causes rather than looking at one-off effects," with a focus on "next-generation technologies, innovations, and nascent industries across sectors."[107] RL's long-term promise, its integral relationship between design choices and domain risks, and its tendency toward unpredictable exo-feedback make it a strong target for future investigations from the FTC. Reward Reports could help structure and define the terms of these investigations, putting designers in a better position to answer agency queries and allowing FTC agents to interrogate specification assumptions comparatively.

**2) Other federal agencies, bureaus, and executive departments that specialize in domain-specific monitoring of harms, standards-setting, and liability.** The case studies of transportation and energy infrastructure that inform this whitepaper would be of particular interest to the Department of Transportation and Environmental Protection Agency, which are charged with protecting these respective domains. Our analysis of the social media setting may also inform prospective initiatives from present or future "AI czars" in the Presidential cabinet,

106 Kratkiewicz, Kendra, Diane Staheli, William Streilein, Dennis Ross, Michael Yee, Olivia Brown, Sanjeev Mohindra, Paul Metzger, and Joseph Zipkin. "Response to the NIST RFI on an Artificial Intelligence Risk Management Framework." Massachusetts Institute of Technology, 2021. URL: https://www.nist.gov/system/files/documents/2021/09/22/ai-rmf-rfi-0089.pdf
107 Khan, Lina. "Vision and Priorities for the FTC". Filed September 22, 2021, available online here: https://www.ftc.gov/system/files/documents/public_statements/1596664/agency_priorities_memo_from_chair_lina_m_khan_9-22-21.pdf

as well as Eric Lander and Alondra Nelson's recent advocacy for an "AI Bill of Rights"[108] at the Office of Science and Technology Policy. We also anticipate the need for oversight from the Food and Drug Administration on RL's application in medical settings. Reward Reports would allow practitioners to make use of more types of feedback from application environments in ways that can be tracked and continuously monitored by regulators of legacy infrastructure.

**3) Civil society organizations that evaluate emergent forms of exo-feedback that are unanticipated by designers.** This includes long-standing organizations like the American Civil Liberties Union or the American Association of People with Disabilities, as well as more recent initiatives that study algorithmic social harms (e.g., Data & Society, the Algorithmic Justice League, AI Now, and the Distributed AI Research Institute). At present, there is a gap between state agencies' ability to monitor domains they are charged with, and the effects anticipated from unrestrained RL optimization. These effects, such as unanticipated carbon emissions or food deserts that emerge from AV-optimized road traffic,[109] may have their roots in behavioral feedback within one domain, but fully manifest in a completely different context. Civil society groups are in a better position to identify and understand the effects of RL systems on distinct subpopulations and provide critical forms of extra-governmental accountability.

Our delineation of feedback types, corresponding risks, and their possible comparative manifestations is intended as a sketch of what more cohesive mappings from reward function to domain may look like. Unlike model cards, the primary value of Reward Reports is not to "accompany a model after careful review has determined that the foreseeable benefits outweigh the foreseeable risks in the model's use or release".[110] Rather, Reward Reports would help make such reviews possible in the first place. In this way, Reward Reports would serve as an intermediate tool that maps reward functions onto emergent behaviors, and evaluates these behaviors against the conditions and features that practitioners anticipate in the deployment setting. Such a tool is necessary not just for the evaluation of future RL systems, but for the governance of automated decision-making systems as a whole.

108 Lander, Eric and Alondra Nelson. "Americans Need a Bill of Rights for an AI-Powered World". Published in *Wired* on October 8, 2021. Available here: https://www.wired.com/story/opinion-bill-of-rights-artificial-intelligence/

109 Igliński, Hubert, and Maciej Babiak. "Analysis of the potential of autonomous vehicles in reducing the emissions of greenhouse gases in road transport." *Procedia engineering* 192 (2017): 353–358.

110 Mitchell, Margaret, Simone Wu, Andrew Zaldivar, Parker Barnes, Lucy Vasserman, Ben Hutchinson, Elena Spitzer, Inioluwa Deborah Raji, and Timnit Gebru. "Model cards for model reporting." In *Proceedings of the conference on fairness, accountability, and transparency*, pp. 220–229. (2019).

# Appendix

## ML DEPLOYMENTS AS RL: THE PROBLEM OF SEQUENTIAL DECISIONS

We have demonstrated that particular characteristics of reinforcement learning give rise to risks distinct from those of supervised machine learning. However, due to forms of feedback that arise when systems are deployed and updated, these risks are relevant to all machine learning deployments, whether they were designed with an RL framework in mind or not.

At present, many domains do not directly apply RL algorithms to problems of *sequential decision-making*, problems in which a decision rule is used repeatedly over time. In the physical world, traditional engineering design still has the upper hand. For example, at Boston Dynamics, the advanced autonomy capabilities of legged robots are achieved through careful modeling and design by control engineers rather than learned from data.[111] In the digital world, there are a variety of challenges that preclude straightforward application of RL techniques: effectively-infinite action spaces (e.g., millions of videos on YouTube), poorly understood dynamics (e.g., varied human behaviors depend on factors external to what a website or app can measure), and the computational impracticality of behavioral feedback (e.g., large models can be re-trained at most daily). Instead, many present-day applications rely on a combination of supervised learning, a collection of occasionally updated models predicting relevant quantities, and hand-designed policies that use these predictions.

There are other frameworks for sequential decision-making that differ in their assumptions of the environment, agent, and reward. These frameworks, when viewed under the lens of reinforcement learning, differ in which types of feedback are incorporated, whether the broader goals are encoded, and how long the relevant time horizon is. In what follows, we explain how these simplified frameworks cast light on particular aspects of RL while eschewing others. By studying the layers of feedback that are elevated or ignored, we illustrate how these frameworks exacerbate or eliminate particular risks when applied to example domains. We also highlight how Reward Reports, when applied broadly to systems conceived within these alternate frameworks, can elucidate and justify the design decisions for particular domains and their risks.

111 Ackerman, Evan. "How Boston Dynamics Taught Its Robots to Dance," IEEE Spectrum, Jan. 7, 2021. URL: https://spectrum.ieee.org/how-boston-dynamics-taught-its-robots-to-dance

## Learning Models in Real Time: Online and Adaptive

One sequential problem that arises in ML deployments is learning in real time from incoming data. Like the supervised learning setting, the overarching goal is to learn an accurate model.

The *online learning* setting[112] considers an agent and a prediction problem. At each timestep, the agent receives a new data point, and must make a prediction. After the prediction is made, the true label of the data point is revealed and the agent can update its model. The performance of the agent is judged by the accuracy of its predictions, and the goal is to quickly learn from early examples so that accurate predictions can be made for incoming examples. For example, a video recommendation site may want to quickly learn a model of user preferences based on browsing behavior.

This setting can be formulated as a special case of reinforcement learning by assigning data points as states, predictions as inputs, the model as the policy, and rewards as predictive accuracy. The system "dynamics" would simply be the data-generating process, which is not modelled as depending on predictions in a meaningful way. Because of this, "actions" do not have an impact on the system's future behavior, and therefore online learning does not consider control feedback or long time horizons. It only considers behavioral feedback in a weak and unidirectional manner: the focus is on how to learn from the incoming data, but not on how to collect more of it in an active manner.

*Active learning*[113] is another framework for sequential decision-making that precisely considers this problem of actively collecting data. It is similar to the online learning problem, except that rather than data points simply appearing, they can be queried by the agent. And rather than accuracy on these queried data points, the goal is to learn a model that will have good predictive accuracy on some larger population. In the video recommendation example, this corresponds to actively choosing videos to suggest, rather than just passively observing user behavior.

Similar to online learning, this setting can be formulated as a special case of reinforcement learning by assigning data points as states, predictions and queries as actions, the model and the rule for querying new data as the policy, and the rewards as accuracy. Now the dynamics are a query model from some underlying data distribution. Actions have a clear impact on the

112 Shalev-Shwartz, S. (2011). Online learning and online convex optimization. *Foundations and trends in Machine Learning*, *4*(2), 107–194.

113 Settles, Burr. "Active learning literature survey." (2009). CS Technical Reports, University of Wisconsin, Madison Computer Sciences Department (2009). URL: http://digital.library.wisc.edu/1793/60660

data observed by the agent, but they are not modelled as having meaningful impact on the underlying distribution. Therefore, while active learning does consider two-way behavioral feedback, it does not consider control feedback.

Active and online learning are focused on prediction and estimation. Broader ML deployments translate these predictions into actions or decisions: "Which ad should be displayed next?" turns into "Which ad is more likely to receive a click?" The decision of how to translate a predicted quantity into an action is made by designers, and the presence of this oversight can reduce risks like reward hacking, especially if the deployment is continually monitored. However, by ignoring control feedback, these paradigms render all of the environment's dynamics as exo-feedback, since they cannot be adequately reasoned about. Instead, they are often dealt with in an ad hoc manner. For example, software engineers resort to a variety of heuristics around notions of novelty, diversity, and serendipity[114] to prevent recommendations from becoming boring and repetitive, since preference models do not incorporate the dynamics of human interest. When problems that arise due to exo-feedback are less immediately obvious to human designers, the system becomes susceptible to Goodhart's Law. As a classic example, once "click prediction" became a dominant factor in social media news feeds, clicks became a worse proxy for valuable content as the metric was gamed by clickbait.

By asking that designers weigh in on the potential dynamics of a domain, Reward Reports can justify the use of learned and continually updated models in deployed automated decision-making systems. The Reward Report will specifically need to include a discussion of why ignoring the effect of decisions on the environment is called for, and how this technical limitation will be supplemented by a plan for monitoring the system's behavior that is commensurate with the stakes of the application. This plan could include metrics as well as mechanisms for interpreting qualitative feedback from users or other affected parties.

## Learning in the Presence of Feedback: Repeated Retraining

Performative prediction[115] is a recently proposed setting that models control feedback directly, motivated by phenomena like Goodhart's Law. Its focus is still on learning a model: an agent seeks to select a parameter that maximizes accuracy over some population. However, the population's distribution is modelled as depending on the chosen parameter. For example, the

114 Herlocker, Jonathan L., Joseph A Konstan, Loren G Terveen, John T. Riedl. "Evaluating collaborative filtering recommender systems," ACM Transactions on Information Systems, vol. 22, no. 1, pp. 5–53 (January 2004). https://doi.org/10.1145/963770.963772

115 Perdomo, Juan, et al. "Performative prediction." *International Conference on Machine Learning*. PMLR, (2020).

presence of a well-known celebrity in a video's thumbnail may initially be highly predictive of clicks, but once a recommender using this model is deployed, the disingenuous use of such thumbnails in obviously low-quality videos will reduce the correlation. Therefore, the goal is to consider parameters that increase accuracy over the distribution that they will induce. A simple but potentially effective approach is *repeated retraining*, where the parameters are repeatedly updated by risk minimization after observing the new distribution.

To relate this setting to reinforcement learning, we assign the population distribution as states, the model parameter as actions, the rule for updating the parameter as the policy, and the expected loss over the new distribution as the reward. The dynamics are the performative rule by which the population updates in response to the parameter. Because of this dependence, this setting models a very strong form of control feedback, where the full effect of the action is immediately realized. Thus a long time horizon is not the focus, as the effects of actions are observed in a single timestep. The performative prediction setting also considers behavioral feedback by focusing on how parameters are updated in response to the changing distribution, e.g., through repeated retraining.

Repeated training broadens the potential risks of many machine learning systems. Although early work on performative prediction characterizes idealized settings where repeated retraining can be effective, the repetition of a process designed for static systems can result in complex and unwieldy dynamics in general. Furthermore, success in the performative prediction setting amounts to ensuring that the population is highly predictable. There is a risk of reward hacking in domains where other goals may not align with predictability, and designers should tread carefully, potentially modifying the loss function to account for this and documenting the reasons for their decisions. Lastly, the relationship between the learned parameter and the underlying distribution implicitly relies on other features of the automated system — for example, how the predictions are used to make a decision that would induce some sort of strategic behavior. If such decision rules are changed, the logic of performativity changes as well. This makes clear the need for reporting and documentation that is specific to each deployment.

## Learning a Policy in Real Time: Explore-Exploit

While the previous discussion considers problems that arise when deploying machine learning systems, they consider performance only through the lens of predictive accuracy. In other words, the primary goal is to ensure that models are accurate. Models are usually a means to some end, however, so there is some other notion of performance that is relevant. In our video recommendation example, accurately identifying a user's preferences over all videos is less important than being able to surface interesting and useful recommendations. Making recommendations is an action, rather than just a prediction. This brings us into the realm of "bandit" problems, which are in some sense an extension of active learning to general rewards. These problems are often motivated with a hypothetical gambling scenario: faced with a slot machine that has multiple arms to pull (the so-called "multi-armed bandit"), how does one efficiently determine which arm will yield the highest winnings? More generally, how should a system sample actions so that it finds the one that achieves the highest overall reward?

In the setting of *contextual bandits*,[116] at each timestep, an agent receives a context and must choose an action. Based on the context and the action, the agent receives some reward. The goal is to design a policy (a map from context to action) that receives high reward. This framework is widely used in targeted advertising applications.

A contextual bandit is often considered by practitioners to be an instance of reinforcement learning.[117] as it includes actions, rewards, and a policy. As far as the designer is concerned, it only remains to assign contexts as states. The dynamics are the data-generating process of contexts, which is not modelled to meaningfully depend on actions. Therefore, like online and active learning, this setting does not consider control feedback. It does consider behavioral feedback, and bandits are the canonical simple setting for understanding the explore-exploit trade-off. In order to receive high reward in the long term, the agent must choose actions that adequately explore the possibilities.

Success in the setting of contextual bandits requires reasoning over long time horizons insofar as the effects of various actions are explored. However, it is not necessary to plan with regards to affecting the underlying environment. As a result, the algorithms that arise from this setting are not optimized to be manipulative. In human-facing applications like video recommendations, these approaches avoid normative questions around modeling human behavioral dynamics and intents. As the goal is to maximize reward, reward hacking remains a risk; however, it is mediated by the fact that algorithms in this setting lack the capacity to plan to manipulate over longer time horizons. However, by ignoring control feedback, the environment's dynamics are offloaded as exo-feedback. By asking designers to hypothesize about underlying dynamics, Reward Reports draw attention to this gap. As in the case for online and adaptive learning, documentation can describe a plan for monitoring the system's behavior to compensate for ignoring control feedback.

116 Chu, Wei, et al. "Contextual bandits with linear payoff functions." *Proceedings of the Fourteenth International Conference on Artificial Intelligence and Statistics*. JMLR Workshop and Conference Proceedings, 2011.
117 See Langford, John et al. "Contextual Bandits," Vowpal Wabbit, 2021. URL: https://vowpalwabbit.org/tutorials/contextual_bandits.html

## Learning a Policy from Offline Data

Offline reinforcement learning is centered around the question: *given a large dataset, can we learn a useful policy without further interactions with the environment?*[118] Offline RL is a variant of the general RL framework that has removed behavioral feedback with the environment. While missing this crucial feedback component may complicate understanding the fundamentals of learning a policy, offline RL is an exciting framework for using multitudes of past data on hand for learning intelligent behavior. This enables it to potentially better leverage fundamental progress in areas like deep learning. It also makes it applicable to domains where trial-and-error strategies online are undesirable for reasons of safety.

Offline RL is also called *fully batch RL*. Here, the data for policy design arrives in a single batch, and the policy is designed in a single shot. At the other end of the spectrum is (*vanilla*) *online RL*, where the agent learns from the data it collects during operation and the policy is constantly updated. Studying fully- or partially-batch RL allows us to separate the effects of sequential decision-making from behavioral feedback. A recent basic research paper proposes a framework, called "Collect & Infer" (C&I), that can be used to study these differences.[119] C&I acts in the conceptual middle of offline RL and vanilla RL by performing updates less frequently, delaying feedback in the behavior loop, while using offline RL algorithms to extract the maximum knowledge from an existing dataset.

Removing interactions during the trial-and-error phase of learning can substantially alter the potential safety of an RL system. Avoiding "errors" is desirable, but without "trial" feedback, it is not clear how the system will achieve its full potential. It is likely that companies will begin with offline RL because it is lower risk than full trial-and-error learning. By modifying the prevalence of behavioral feedback, offline RL concentrates its potential risks. Without the capability to explore the environment and collect its own data, an offline RL policy is less prone to reward hacking. The extent to which reward hacking can occur can be understood from the offline dataset used for training. In some sense, documentation is easier since it can focus on the single deployed policy and the offline dataset used to generate it. However, without the ability to learn online, concerns of exogenous feedback and environmental drift become more prevalent. Documentation should thus still include a plan for continuous monitoring that is sensitive to mismatch between the past data and the current world.

118 Levine, Sergey, et al. "Offline reinforcement learning: Tutorial, review, and perspectives on open problems." *arXiv preprint arXiv:2005.01643.* (2020).
119 Riedmiller, Martin, et al. "Collect & Infer--a fresh look at data-efficient Reinforcement Learning." *arXiv preprint arXiv:2108.10273.* (2021).

# Acknowledgments

For their helpful comments and suggestions on this paper, the authors wish to thank Daniel Aranki, Eugene Bagdasaryan, Anthony Barrett, Jared Brown, Micah Carroll, Lucia Cipolina-Kun, Roel Dobbe, Blake Elias, Dylan Hadfield-Menell, Emmanuel Moss, Jessica Newman, Brandie Nonnecke, Frank Pasquale, Aaron Snoswell, and Eugene Vinitsky. The authors also wish to thank the Center for Long-Term Cybersecurity and the Center for Human-Compatible AI for supporting previous stages of research resulting in this paper.

# About the Authors

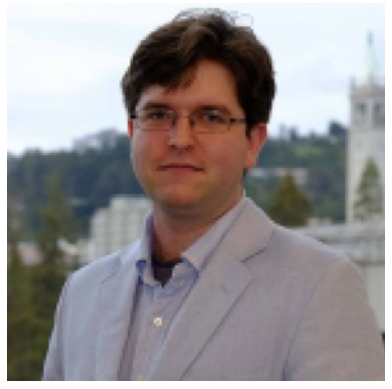

**Thomas Krendl Gilbert** is a postdoctoral fellow at the Digital Life Initiative at Cornell Tech in New York City. He previously designed and received a Ph.D. in Machine Ethics and Epistemology at the University of California at Berkeley, Berkeley, CA, USA. Thomas researches the emerging political economy of AI, in particular reinforcement learning systems.

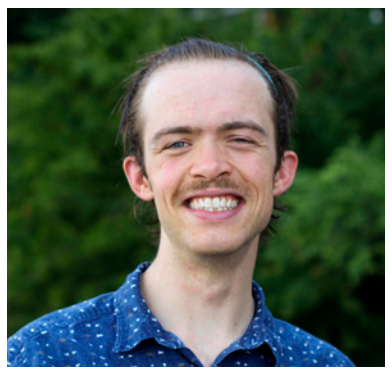

**Nathan Lambert** received the B.S. degree in electrical and computer engineering from Cornell University, Ithaca, NY, USA, in 2017. He is currently pursuing the Ph.D. degree with the University of California at Berkeley, Berkeley, CA, USA. He is a member of the Department of Electrical Engineering and Computer Sciences, advised by Prof. K. Pister with the Berkeley Autonomous Microsystems Lab. His work explores many topics on model learning and decision-making with data-driven and analytical methods.

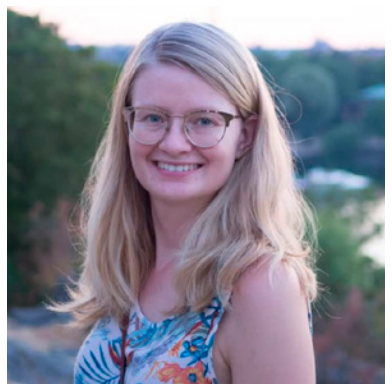

**Sarah Dean** is an Assistant Professor in the Computer Science Department at Cornell University, Ithaca, NY. She completed her Ph.D. in the Department of Electrical Engineering and Computer Science at University of California at Berkeley in 2021. Her research focuses on developing principled data-driven methods for control and decision-making, inspired by applications in robotics, recommendation systems, and developmental economics.

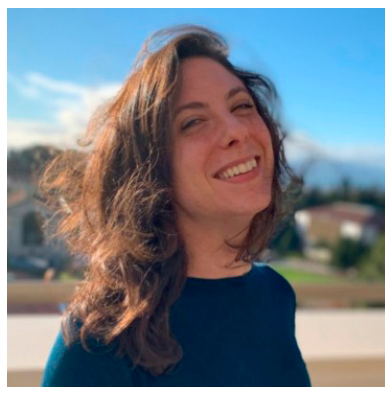

**Tom Zick** is a researcher in AI ethics at the Berkman Klein Center for Internet and Society at Harvard University, where she is also a J.D. candidate. She holds a Ph.D. from UC Berkeley and was previously a fellow at Bloomberg Beta and the City of Boston. Her research centers on legal mechanisms for oversight of machine learning systems, particularly in the context of reinforcement learning systems.

CLTC

Center for Long-Term Cybersecurity

UC Berkeley

Center for Long-Term Cybersecurity
cltc.berkeley.edu
@CLTCBerkeley